\documentclass[pdflatex,sn-mathphys-num]{sn-jnl}

\usepackage{graphicx}%
\usepackage{multirow}%
\usepackage{amsmath,amssymb,amsfonts}%
\usepackage{amsthm}%
\usepackage{mathrsfs}%
\usepackage[title]{appendix}%
\usepackage{xcolor}%
\usepackage{textcomp}%
\usepackage{manyfoot}%
\usepackage{booktabs}%
\usepackage{algorithm}%
\usepackage{algorithmicx}%
\usepackage{algpseudocode}%
\usepackage{listings}%

\usepackage{adjustbox}
\usepackage{CJKutf8}
\usepackage{float}
\usepackage{lscape}
\usepackage{multirow}
\usepackage{makecell}

\theoremstyle{thmstyleone}%
\theoremstyle{thmstyletwo}%

\theoremstyle{thmstylethree}%

\begin{document}

\title[Article Title]{Domain Adaptation for Japanese Sentence Embeddings with Contrastive Learning based on Synthetic Sentence Generation}


\author*[1]{\fnm{Zihao} \sur{Chen}}\email{cyjk2101@mail4.doshisha.ac.jp}

\author[2]{\fnm{Hisashi} \sur{Handa}}\email{handa@info.kindai.ac.jp}

\author[1]{\fnm{Miho} \sur{Ohsaki}}\email{mohsaki@mail.doshisha.ac.jp}

\author[1]{\fnm{Kimiaki} \sur{Shirahama}}\email{kshiraha@mail.doshisha.ac.jp}

\affil[1]{\orgdiv{Graduate School of Science and Engineering}, \orgname{Doshisha University}, \orgaddress{\street{1-3, Tatara Miyakodani}, \city{Kyotanabe}, \postcode{610-0394}, \state{Kyoto}, \country{Japan}}}

\affil[2]{\orgdiv{Faculty of Informatics}, \orgname{Kindai University}, \orgaddress{\street{3-4-1, Kowakae}, \city{Higashiosaka}, \postcode{577-8502}, \state{Osaka}, \country{Japan}}}



\abstract{Several backbone models pre-trained on general domain datasets can encode a sentence into a widely useful embedding. Such sentence embeddings can be further enhanced by domain adaptation that adapts a backbone model to a specific domain. However, domain adaptation for low-resource languages like Japanese is often difficult due to the scarcity of large-scale labeled datasets. To overcome this, this paper introduces SDJC (Self-supervised Domain adaptation for Japanese sentence embeddings with Contrastive learning) that utilizes a data generator to generate sentences, which have the same syntactic structure to a sentence in an unlabeled specific domain corpus but convey different semantic meanings. Generated sentences are then used to boost contrastive learning that adapts a backbone model to accurately discriminate sentences in the specific domain. In addition, the components of SDJC like a backbone model and a method to adapt it need to be carefully selected, but no benchmark dataset is available for Japanese. Thus, a comprehensive Japanese STS (Semantic Textual Similarity) benchmark dataset is constructed by combining datasets machine-translated from English with existing datasets. The experimental results validates the effectiveness of SDJC on two domain-specific downstream tasks as well as the usefulness of the constructed dataset. Datasets, codes and backbone models adapted by SDJC are available on our github repository \url{https://github.com/ccilab-doshisha/SDJC}.}




\maketitle

\section{Introduction}
\label{sec:sec1}



Sentence representation learning is a fundamental problem in Natural Language Processing (NLP). It learns to encode sentences into compact, dense vectors, called \textit{embeddings}, which can be broadly useful for a wide range of tasks, such as \textit{Semantic Textual Similarity} (STS)~\cite{zhang2020unsupervised}, language understanding and evaluation~\cite{cer2018universal, perone2018evaluation}, and information retrieval~\cite{thakur2021beir,wang2022just}. However, since most language models are pre-trained on a general domain, sentence embeddings directly generated from them are unsatisfactory for a specific domain. Recent work has shown the importance of \textit{domain adaptation} that adapts or fine-tunes a pre-trained model to the specific domain~\cite{schopf2023efficient,gururangan2020don,guo2021multireqa,ke2022continual}. The pre-trained model is supposed to produce embeddings that capture common characteristics across various domains. Domain adaptation enables it to offer higher-quality embeddings for the specific domain, compared to embeddings obtained by a model that is trained from scratch~\cite{sennrich2015improving,lee2020biobert}. In what follows, a pre-trained language model to be adapted is called a \textit{backbone model}.

One promising approach to domain adaption is \textit{contrastive learning} which adapts a backbone model to a specific domain so that embeddings of semantically similar sentence pairs are close to each other while those of dissimilar pairs are pushed apart~\cite{chen2022generate, wang2020understanding, wang2021understanding}. The former type of pairs and the latter type of pairs are called \textit{positive pairs} and \textit{negative pairs}, respectively. The distinction between positive and negative pairs enables the adapted backbone model to output embeddings capturing semantic concepts and their relations in the specific domain. Existing domain adaptation methods based on contrastive learning create sentence pairs in a supervised or unsupervised way. Supervised methods apply domain-specific human-annotated sentence pairs. Meanwhile, unsupervised methods leverage unlabeled corpora and generate synthesized sentence pairs or pseudo labels. In general, supervised methods outperform unsupervised ones although the former need large-scale labeled training datasets which are usually not available.

Languages can be divided into \textit{high-resource} and \textit{low-resource}. High-resource languages like English are characterized by the fact that many large-scale labeled datasets are available. Thus, it is not so difficult to find a large-scale labeled dataset that is suitable for a domain-specific downstream task. Or, such a dataset may be prepared by slightly editing existing datasets. On the other hand, low-resource languages like Japanese mean that only a limited number of datasets are available. It is highly likely that no appropriate dataset is available for a domain-specific downstream task. Moreover, editing an existing dataset is often useless because domains of available datasets are different from the one of the downstream task.

To alleviate the lack of large-scale labeled datasets for a low-resource language, this paper introduces a \textit{self-supervised} domain adaptation approach that generates synthetic sentences describing semantically different meanings from available sentences. The approach leverages contrastive learning to adapt a backbone model by contrasting original sentences with their synthetic counterparts. Since this paper especially targets at Japanese, our approach is named SDJC which is derived from ``Self-supervised Domain adaptation for Japanese sentence embeddings with Contrastive learning''. To begin with, SDJC needs a domain-specific unlabeled corpus collected from a target domain. SDJC then fine-tunes a pre-trained data generator (in our case T5 (Text-to-Text Transfer Transformer)~\cite{raffel2020exploring}) on the collected corpus using unsupervised training with a fill-in-the-blank-style denoising objective. Based on our preliminary experiment in Section~\ref{ssec:relevant_content_words}, the fine-tuned data generator is used to substitute nouns in an input sentence to construct a semantically dissimilar one. That is, the input sentence and the constructed one form a negative pair. On the other hand, by borrowing the idea in \cite{gao2021simcse}, a positive pair is formed on the embedding level. The embedding obtained for the input sentence using a backbone model is paired with an embedding that is for the same input sentence but is obtained by randomly dropping units of the backbone model. Finally, SDJC adapts the backbone model by performing contrastive learning with domain-specific positive and negative pairs.

Implementing SDJC requires careful selections for its components. Two components that are strongly linked to SDJC's performance are a proper sentence representation learning method and a backbone model to be adapted. For the former, several existing methods have been proposed, but the question is which one best suits SDJC. The latter means selecting the backbone model that leads to the best performance after SDJC's domain adaptation. However, these selections involve a problem that is also attributed to a limited number of available datasets for a low-resource language. Namely, no benchmark dataset is available to evaluate performances with different combinations of a sentence representation learning method and a backbone model in Japanese.

Taking English as an example of high-resource language, previous research usually follows the evaluation protocol that uses SentEval toolkit~\cite{conneau2018senteval} on the standard STS benchmark. Accordingly, several datasets for STS tasks have been released by targeting English. 
Researchers have also released datasets for \textit{Natural Language Inference} (NLI) which is a task to predict the relationship between a pair of sentences, such as entailment, neutral or contradiction.
For Japanese as an example of low-resource language, only JSNLI (Japanese version of the Stanford NLI)~\cite{yoshikoshi2020multilingualization} and JSICK (Japanese version of Sentences Involving Compositional Knowledge)~\cite{yanaka2022compositional} datasets are publicly available. JSNLI dataset is a translated version of SNLI~\cite{bowman2015large} dataset using machine translation, and JSICK dataset is made manually based on SICK~\cite{marelli-etal-2014-sick} dataset. Datasets for many other tasks like STS12-16~\cite{agirre2012semeval, agirre2013sem, agirre2014semeval, agirre2015semeval, agirre2016semeval}, STS-B~\cite{cer2017semeval} and 7 transfer tasks from SentEval toolkit~\cite{conneau2018senteval} are unavailable for Japanese. Moreover, existing contrastive learning methods have so far been tested mainly on English sentences. It is unknown whether their performance comparison for English is also valid for Japanese.


Therefore, we construct a comprehensive Japanese STS (JSTS) benchmark dataset including multiple STS datasets in order to examine the generalities of existing sentence representation learning methods. Referring to previous research~\cite{yoshikoshi2020multilingualization, ham2020kornli}, machine translation is used to translate English STS12-16 and STS-B datasets into Japanese. Our comprehensive JSTS benchmark dataset is built by combining the translated datasets with JSICK~\cite{yanaka2022compositional} dataset and Japanese STS (JSTS) dataset which is a part of JGLUE (Japanese General Language Understanding Evaluation) benchmark~\cite{kurihara2022jglue}. The experimental results using our benchmark dataset reveal that existing contrastive learning methods which have been validated for English so far, work well also for Japanese. In other words, contrastive learning yields performance improvements of backbone models for Japanese. Furthermore, the experimental results allow us to choose the contrastive learning method in \cite{gao2021simcse} and BERT~\cite{devlin2018bert} as the sentence representation learning method and the backbone model of SDJC, respectively.

Finally, SDJC is tested on two domain-specific downstream tasks, one is an STS task of Japanese sentences in a clinical domain using Japanese Clinical STS (JACSTS) dataset~\cite{mutinda2021semantic}, and the other is an information retrieval task in an educational domain using QABot dataset that consists of questions submitted by students via a messaging app (Slack) in a programming exercise course~\cite{chen2022retrieval,ide2023qabot}. The results show that SDJC's domain adaptation based on contrastive learning with its generated positive and negative pairs achieves superior performances in comparison to no domain-specific adaptation. It is also demonstrated that further performance improvements are attained by additional fine-tuning on JSNLI or Wikipedia dataset.

To sum up, the main contributions of this paper are summarized as follows:
\begin{itemize}
    \item We propose SDJC that is a novel, effective domain adaptation approach for Japanese sentence embeddings. SDJC adapts a backbone model to a domain-specific downstream task by contrastive learning on automatically generated sentence pairs.
    \item A comprehensive JSTS benchmark dataset is constructed. To our best knowledge, this paper is the first to benchmark several sentence representation learning methods and backbone models on Japanese sentences.
    \item On our github repository \url{https://github.com/ccilab-doshisha/SDJC}, we release our comprehensive JSTS benchmark dataset as well as SDJC's codes, adapted backbone models and datasets that are free from the ethical issue. We believe that the former is useful for many other purposes than selecting a sentence representation method and a backbone model of SDJC in this paper, and the latter is also valuable for devising various applications based on high-quality Japanese sentence embeddings.
\end{itemize}

Finally, this paper significantly differs from \cite{chen2022retrieval,chen2023domain,chen2023jcse} that we published in the middle of completing this paper. First, \cite{chen2022retrieval} is a short paper presenting our initial result of the information retrieval task on QABot dataset. Here, a pre-trained model (Sentence-BERT) was used to compute the embedding of each question with no domain adaptation. Second, \cite{chen2023domain} is a one-page paper showing our initial insights into domain adaptation by contrastive learning using automatically generated sentence pairs, and \cite{chen2023jcse} is a long version of \cite{chen2023domain}. However, these papers have the following two big problems: The one is that our main contributions are unclear and the other is insufficient experiments. To resolve them, we form this paper by drastically revising \cite{chen2023jcse}, more specifically, revising almost all the sentences as well as the organization of sections, reviewing much more related studies, obtaining new results with different backbone models and performing visualizations of sentence embeddings.

\section{Related Work}
\label{sec:related_work}

This section starts with reviewing various sentence embedding methods and describing the difference between them and SDJC. Afterwards, the novelties of SDJC in terms of contrastive learning and domain adaptation are discussed by checking existing methods in these fields. Lastly, the uniqueness of our Japanese STS benchmark dataset is clarified together with the discussion of methods for constructing non-English datasets.

\subsection{Sentence Embedding}

In the context of machine learning, embeddings make it easier to process large size inputs such as audio, image and text. Ideally, an embedding captures some semantics of an input by placing semantically similar inputs close together in a lower-dimensional space. In NLP, an embedding is defined as a numerical representation of a word, sentence or document in such a way that the representation numerically conserves the meaning of the target~\cite{pilehvar2022embeddings, li2022brief}. Modern NLP systems rely on embeddings pre-trained on large corpora as base features.

Existing sentence embedding methods can be divided into two types named \textit{word-view} and \textit{sentence-view}~\cite{li2022brief}. The former forms an embedding of a sentence from words in a seemingly simple way such as weighted summation, concatenation, or output of a neural network. These methods are computationally efficient but underperforming~\cite{wieting2015towards, arora2017simple}.

The sentence-view type emphasizes the interaction between sentences, ranging from different training objectives with diverse neural network architectures.
Skip-thought~\cite{kiros2015skip} extends the context-dependency idea behind the skip-gram model to the sentence level by utilizing a central sentence to predict adjacent sentences using GRUs.
Moreover, InferSent~\cite{conneau2017supervised} performs the NLI task using multiple designed sentence encoder architectures based on CNN, GRU, and LSTM networks~\cite{liu2016learning, lin2017structured}. While the aforementioned methods employ a single training objective, several hybrid multi-task approaches optimize multiple objectives by jointly addressing sentence coherence and semantics. 
Subramanian \textit{et al.}~\cite{subramanian2018learning} proposed a simple but effective multi-task learning framework including Skip-thought's adjacent sentence prediction, neural machine translation, constituency parsing~\cite{vinyals2015grammar} and NLI tasks to train a shared bi-directional GRU encoding model. USE~\cite{cer2018universal} follows the same multi-task idea but substitutes the model with the encoder part of the transformer architecture~\cite{Transformers}, highlighting the significant potential of this new architecture. However, the above-mentioned methods train models from scratch without using any pre-trained models, which requires a huge computational cost.


In the sentence-view type, researchers have recently developed methods that leverage pre-trained transformer-based language models like BERT~\cite{devlin2018bert}. The original BERT shows an unsatisfactory performance on sentence embeddings \cite{li2020sentence, su2021whitening}. One explanation for this is anisotropy which means that embeddings occupy a narrow cone in the vector space where embeddings of two irrelevant sentences are unexpectedly close to each other in terms of angular distance~\cite{li2020sentence, godey2024anisotropy}. Some work \cite{li2020sentence, su2021whitening} focused on reducing the anisotropy by post-processing sentence embeddings obtained by a pre-trained BERT. 
Sentence-BERT~\cite{reimers-gurevych-2019-sentence} applies siamese and triplet networks to fine-tune a pre-trained BERT on NLI dataset.
Gao \textit{et al.}~\cite{gao2021simcse} utilized contrastive learning to help a pre-trained BERT learn better sentence embeddings. 
PromptBERT~\cite{jiang2022promptbert} proposed a prompt-based contrastive learning method where a sentence is incorporated into a designed prompt (template) to represent a better embedding from a pre-trained BERT. 
GenSE~\cite{chen2022generate} proposed an improved prompt-based contrastive learning method where designed prompts are fed into a trained generative model to generate useful positive and negative pairs.
It has recently shown that Large Language Models (LLMs) pre-trained on massive datasets achieve impressive results in various NLP tasks with no fine-tuning. 
PromptEOL~\cite{jiang-etal-2024-scaling} employs in-context learning where an elaborately designed prompt with created demonstration sentence and word pairs is fed into an LLM to obtain sentence embeddings that lead to comparable performances with current state-of-the-art contrastive learning methods.


However, the above sentence-view type methods using pre-trained models need human-annotated datasets~\cite{li2020sentence, su2021whitening,reimers-gurevych-2019-sentence,gao2021simcse,jiang2022promptbert,chen2022generate} or careful manual engineering of prompts~\cite{chen2022generate, jiang2022promptbert,jiang-etal-2024-scaling}. Especially, human-annotated or even domain-specific datasets are not expected for low-resource languages like Japanese. Thus, SDJC adopts an approach to leverage a raw corpus that can be easily collected. Then, this raw corpus is used to train a data generator to generate negative sentence pairs, unlike GenSE~\cite{chen2022generate} that trains a generator using a human-annotated dataset. The previous work~\cite{jiang2022promptbert, zhang2022unsupervised, chuang2022diffcse, gao2021simcse} applying contrastive learning also adapts unsupervised paradigms leveraging large-scale unlabeled datasets although the performance is still behind the supervised counterparts.

\subsection{Contrastive Learning}
\label{ssec:contrastive_learning}

In contrastive learning, a sentence called \textit{anchor sentence} is compared to two types of sentences, \textit{positive sentence} and \textit{negative sentence}, that are semantically similar and dissimilar to the anchor sentence, respectively. Thus, the embedding of the anchor sentence and the one of the positive sentence form a positive pair, which indicates that these embeddings should be similar to each other. On the other hand, the embedding of the anchor sentence and the one of the negative sentence define a negative pair signifying that they should be different from each other.

One crucial question in unsupervised contrastive learning is how to construct positive and negative pairs. Some research~\cite{yan2021consert, wu2020clear} prepare positive sentences by applying multiple augmentation techniques such as word deletion, reordering, cutoff, and synonym substitution to anchor sentences. Kim \textit{et al.}~\cite{kim2021self} make a positive pair by uniform sampling of hidden layers of a pre-trained BERT. That is,  the embedding of an anchor sentence is paired with the embedding of a sampled layer. SimCSE~\cite{gao2021simcse} simply feeds the same anchor sentence to a model twice using random unit dropout of BERT and constructs a positive pair by pairing the resulting two different embeddings. 
Previous methods mainly elaborate on how to construct positive pairs while the construction of negative pairs lacks a meticulous design and follows the standard practice to pair an anchor sentence with another sentence within a minibatch by assuming that they convey different meanings~\cite{yan2021consert, wu2020clear, gao2021simcse}. 
However, explicit negative pairs also play a crucial role.
Several studies~\cite{wang2021understanding, xuan2020hard, khosla2020supervised} in the computer vision community show the importance of \textit{hard negative pairs} each of which is defined as a pair of features (embeddings) that are similar but derived from semantically dissimilar examples (in our case sentences).
Robinson \textit{et al.}~\cite{robinson2020contrastive} employs importance sampling to reweight the contrastive learning objective function by a novel sampling distribution that weights the negatives in a minibatch based on their similarities to the anchor example.
For NLP, MixCSE~\cite{zhang2022unsupervised} constructs an artificial negative embedding by mixing the embedding of a positive sentence and the one of a random negative sentence. 
SNCSE~\cite{wang2023sncse} focuses on applying pre-defined rules to generate negated versions of sentences as soft negative sentences.
Zhuang \textit{et al.}~\cite{zhuang2024not} adapts a similar idea with Robinson \textit{et al.}~\cite{robinson2020contrastive} that individualizes either the weight or temperature of each negative pair according to the similarities of negative sentences to anchors.

 
Compared to the existing methods described above, we propose a self-supervised construction of hard negative pairs by adopting a data generator that generates a \textit{hard negative sentence} that has the same syntactic structure as an anchor sentence but conveys a different semantic meaning. Previous research~\cite{yanaka2022compositional} examines whether Japanese language models are sensitive to word orders and case particles for the prediction of entailment labels. This inspires us to investigate which word types in Japanese are the most relevant for different Japanese language models. In what follows, we use a \textit{relevant content word} to indicate the word that is the most influential to determine the entailment (similarity) of a sentence pair. Wang \textit{et al.}~\cite{wang2021tsdae} introduce an approach to identify the most relevant content word in English based on how masking each word within a sentence affects its embedding. In Section~\ref{ssec:relevant_content_words}, we extend this approach to Japanese and show that nouns are the most relevant content words in Japanese sentences. Hence, we propose a self-supervised contrastive learning method where sentences in a given corpus are treated as anchor sentences, and hard negative sentences for them are automatically created by substituting nouns in them.



\subsection{Domain Adaptation for Domain-specific Downstream Tasks}
\label{ssec:related_work_domain_adaptation}

The most widely used and straightforward approach to domain adaption in NLP is Domain-Adaptive PreTraining (DAPT), which involves continuing the pretraining of a backbone model on a large raw (unlabeled) domain-specific corpus. 
Gururangan \textit{et al.}~\cite{gururangan2020don} address eight classification tasks ranging across four domains, biomedical publications, computer science publications, newstext and reviews, and show that DAPT consistently improves performances on tasks in each domain. BioBERT~\cite{lee2020biobert} shows that pre-training BERT on large-scale biomedical corpora from scratch largely improves its performance on downstream tasks of biomedical text mining. Meanwhile, Guo \textit{et al.}~\cite{guo2021multireqa} and Ma \textit{et al.}~\cite{ma2021zero} show that neural retrieval models trained on a general domain often do not transfer well to specific domains. Thus, domain adaptation using a domain-specific corpus is essential since even a model of hundreds of millions of parameters pre-trained on general corpora struggles to perform well on domain-specific downstream tasks.
Compared to the above-mentioned work where a domain-specific raw corpus is directly used like DPAT, we leverage such a corpus to automatically generate sentence pairs so that contrastive learning can be performed to fine-tune a backbone model. That is, we combine domain adaptation with sentence representation learning (contrastive learning) in a self-supervised fashion. 


Domain adaptation with sentence representation learning has been explored recently. 
Schopt \textit{et al.}~\cite{schopf2023efficient} propose to adapt sentence embeddings to a specific domain by injecting a lightweight domain-specific adapter into a backbone model. 
AdaSent~\cite{huang2023adasent} assembles DAPT and sentence representation learning with an adapter together to perform few-shot classification using a small amount of labeled data.
However, these methods need labeled domain-specific data that are difficult to collect in some domains.
To overcome this, leveraging generative models to produce synthetic domain-specific data has been widely explored. 
Ma \textit{et al.}~\cite{ma2021zero} and Liang \textit{et al.}~\cite{liang2020embedding} use synthetic queries for domain adaptive neural retrieval.
More specifically, they leverage synthetic queries generated from a large sequence-to-sequence (seq2seq) model based on domain-specific passages like biomedical literature or documents from Ubuntu forums as training data.
Yue~\textit{et al.}~\cite{yue2021cliniqg4qa} leverage a seq2seq-based question phrase prediction module to enable question generation models to synthesize more diverse questions on clinical domains and boost QA models without requiring labeled data.
Recent advancements in generating domain-specific data for sentence representation learning with domain adaptation include SynCSE~\cite{zhang2023contrastive}, which leverages ChatGPT to produce positive and negative sentences for contrastive learning. This process involves carefully designed human-crafted prompts tailored to a specific domain, ensuring that the generated sentences align with the characteristics of the domain.
GenSE~\cite{chen2022generate} jointly trains a generator and a discriminator using a labeled NLI dataset to generate positive and negative sentences with designed prompts for prompt-based contrastive learning on specific domains. These two leading works are not practical for low-resource languages like Japanese, for which extensive labeled data are often unavailable and a well-trained generative model together with carefully designed prompts is needed.

To overcome this, we leverage a lightweight and adaptable framework where specific-domain sentence pairs are generated without requiring labeled data or carefully designed prompts. Specifically, we fine-tune a small-scale generative model (T5~\cite{raffel2020exploring}) with an unlabeled domain-specific corpus to produce high-quality hard negative sentences. This way, our approach is designed to be resource-efficient and applicable to low-resource language settings.

\subsection{Non-English Data Construction}
\label{ssec:non-english_datasets}

With the advent of multilingual pre-trained language models, general language understanding frameworks have been extended to languages beyond English, and NLI datasets have been adapted to multilingual settings.
Conneau \textit{et al.}~\cite{conneau2018xnli} introduced Cross-lingual Natural Language Inference (XNLI) dataset which was created by translating Multi-Genre Natural Language Inference (MNLI) dataset~\cite{williams2018broad} into 15 languages, including low-resource languages such as Urdu and Swahili. Notably, Japanese is not included in XNLI dataset. 
Ham \textit{et al.}~\cite{ham2020kornli} create KorNLI and KorSTS datasets by translating English NLI and STS-B~\cite{cer2017semeval} datasets into Korean. There are also attempts to translate SICK dataset~\cite{marelli-etal-2014-sick} into Dutch~\cite{wijnholds2021sick} and Portuguese~\cite{real2018sick}.
Rather than translating from English datasets, 
OCNLI~\cite{hu2020ocnli} introduced a Chinese NLI dataset built from original Chinese multi-genre resources. 
Similarly, TyDi QA~\cite{clark2020tydi} proposed a multilingual QA dataset built from scratch, using native speakers to generate information-seeking questions directly in 11 diverse languages such as Arabic, Finnish, and Swahili.

Regarding Japanese, JSNLI dataset~\cite{yoshikoshi2020multilingualization} is constructed by employing machine translation to translate English SNLI dataset into Japanese and automatically filtering out unnatural sentences. JSICK~\cite{yanaka2022compositional} dataset is manually translated from English SICK dataset. In contrast, Yahoo Japan developed JGLUE benchmark~\cite{kurihara2022jglue} from scratch without translation due to the cultural/social discrepancy between English and Japanese. Although employing machine translation incurs several possibilities to produce unnatural sentences, it is more cost-efficient than manual translation and reconstruction from scratch.

Our comprehensive JSTS benchmark dataset is the first of its kind for the Japanese STS task. We employ machine translation to translate standard STS datasets in English (i.e., STS12-16 and STS-B) into Japanese.  The translated datasets are then combined with human-annotated datasets (i.e., JSICK~\cite{yanaka2022compositional} and JSTS~\cite{kurihara2022jglue}), in order to examine whether existing sentence representation learning methods generally work on both machine-translated and human-annotated  datasets.

\section{Domain Adaptation for Japanese Sentence Embeddings with Contrastive Learning}
\label{sec:domain_adaption_for_japanese_sentence_embeddings}

\begin{figure}[htbp]
    \centering
    \includegraphics[width=\linewidth]{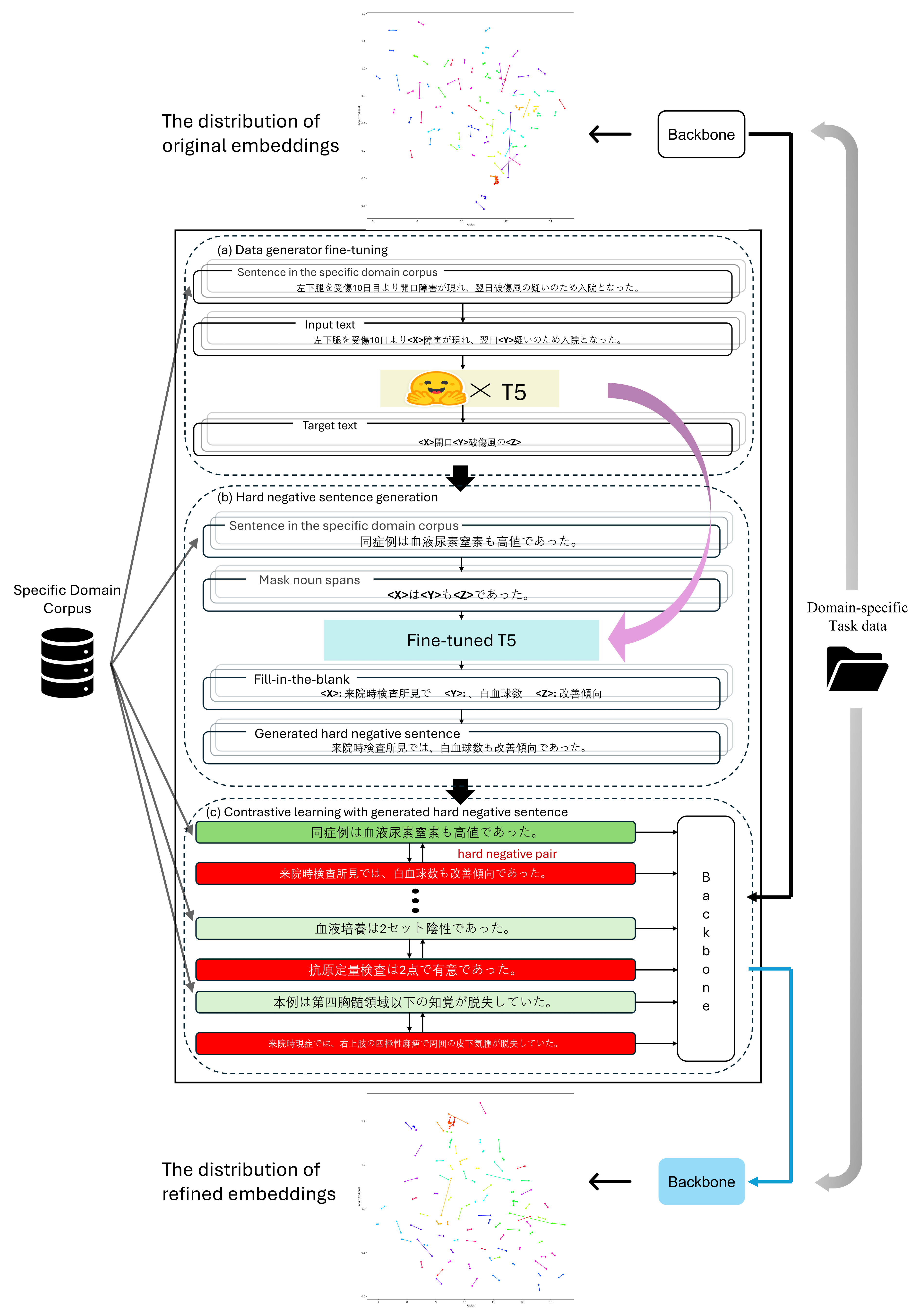}
    \caption{An overview of SDJC.}
    \label{fig1}
\end{figure}

Fig.~\ref{fig1} illustrates an overview of SDJC. First of all, our purpose is to adapt a backbone model to a domain-specific task so that it can output higher-quality embeddings of sentences in the specific domain, as shown in the top and bottom of Fig.~\ref{fig1} (please refer to Section~\ref{ssec:visualization_se} for details of the embedding visualization). To this end, a raw (unlabeled) specific domain corpus shown on the left of Fig.~\ref{fig1} is firstly collected from public resources. As depicted in Fig.~\ref{fig1} (a), a pre-trained data generator (T5~\cite{raffel2020exploring} in this paper) is then fine-tuned using this corpus in order to make it align with the linguistic and contextual properties of the specific domain. 
Afterward, Fig.~\ref{fig1} (b) illustrates that the fine-tuned data generator is employed to generate hard negative sentences by regarding sentences in the specific domain corpus as anchor sentences and replacing nouns in them. Finally, Fig.~\ref{fig1} (c) exhibits that these anchor and hard negative sentences are exploited to perform contrastive learning for adapting the backbone model to the domain-specific task. Thereby, the adapted backbone model can better capture nuanced domain-specific semantic relationships. 
More details of the above-mentioned processes are described below.

\subsection{Collecting target domain corpora}
\label{ssec:specific_domain_corpus_collection}

A backbone model and a data generator employed in SDJC are pre-trained on general-domain corpora without incorporating sufficient knowledge of a specific domain. This generalization results in a significant mismatch between the word distributions of the general and specific domain corpora. Consequently, the pre-trained backbone model struggles to generate high-quality sentence embeddings that effectively capture domain-specific nuances such as technical jargon and specialized linguistic structures. Similarly, the data generator fails to produce sentences that are contextually relevant and appropriate for the specific domain. 
Hence, it is essential to collect a specific domain corpus and adapt the backbone model and the data generator to the specific domain using the collected corpus.
However, imposing manual annotation on this corpus collection is laborious and impractical. Thus, we opt to collect a raw corpus from publicly available resources of the specific domain. In particular, the following two corpora are created by targeting a clinical domain and an educational domain.

For the Japanese clinical domain, the reason for few resources is due to the data privacy which prohibits public sharing of medical data. We collect data from two available sources, Japanese Case Reports (CR) and Japanese NTCIR-13 MedWeb datasets. CR dataset is created by extracting $200$ case reports from CiNii that is a Japanese database containing research publications\footnote{https://cir.nii.ac.jp/}. Japanese NTCIR-13 MedWeb dataset is a part of NTCIR-13 Medical NLP for Web Document (MedWeb) task at NTCIR-13 conference\footnote{http://mednlp.jp/medweb/NTCIR-13/} and provides $2560$ Twitter-like message texts. We combine these two datasets while dropping duplicate sentences.


For the Japanese educational domain, we focus on a further specialized domain, namely the Japanese educational computer science domain, which requires us to treat interdisciplinary characteristics of integrating knowledge about education and computer science. Focusing on courses offered by Faculty of Informatics at Kindai University in Japan, we leverage syllabus data and teaching materials published in Google Classroom. In addition, private question-answer data are collected by running a QABot~\cite{ide2023qabot} on Slack\footnote{https://slack.com/} which is a popular instant messaging software utilized in different university courses. We aggregate data from these three resources (i.e., syllabus, Google Classroom and QABot data).



Each of the aforementioned corpora is cleaned and normalized by the following processes: First, GINZA\footnote{https://github.com/megagonlabs/ginza}, an open-source Japanese NLP library, is used to divide the corpus into sentences. Next, we carry out text normalization like removing markup and non-text content, then drop sentences whose length is less than five after the tokenization by GINZA. 
Table \ref{corpus_stastic} shows the statistics of the collected corpora.
The corpus in the clinical domain is available on \url{https://github.com/ccilab-doshisha/SDJC}, while the release of the corpus in the education computer science domain is restrained due to the ethical issue.

\begin{table}[htbp]
    \centering
    \begin{adjustbox}{width=\textwidth}
    \begin{tabular}{ccccc}
    \hline
    Domain tag & \#Examples & Avg. \#Tokens per Sentence & \#Unique tokens \\
    \midrule
    Clinical & 11832 & 38 & 15806 \\ 
    Educational Computer Science  & 6872 & 52 & 10432\\ \hline        
    \end{tabular}
    \end{adjustbox}
    \caption{Tokens are tokenized by GINZA.}
    \label{corpus_stastic}
\end{table}

\begin{figure}[H]
    \centering
    \includegraphics[width=\textwidth]{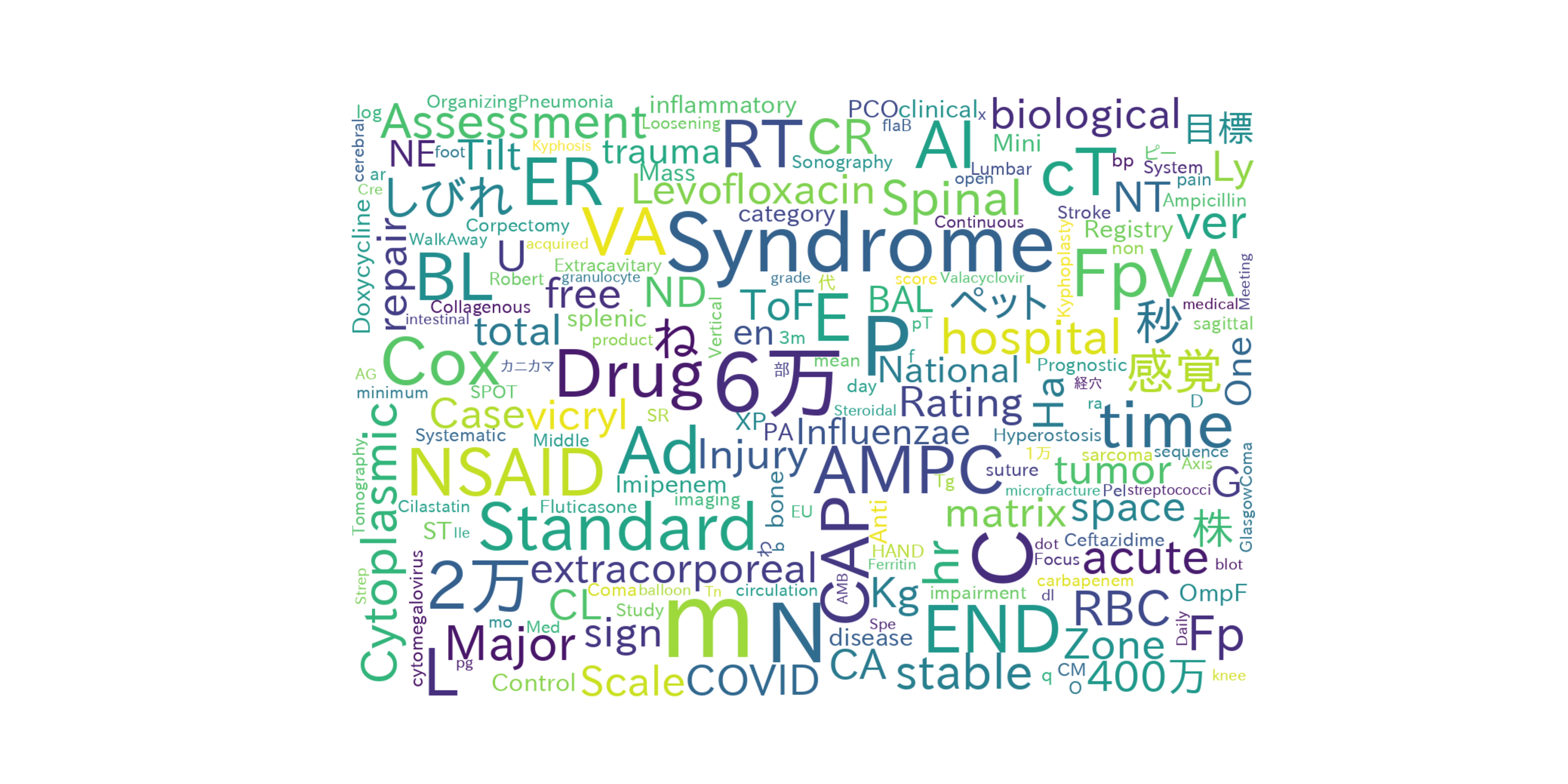}
    \caption{A Word Cloud plot of our collected corpus in the clinical domain.}
    \label{word_cloud_clinic}
\end{figure}

\begin{figure}[H]
    \centering
    \includegraphics[width=\textwidth]{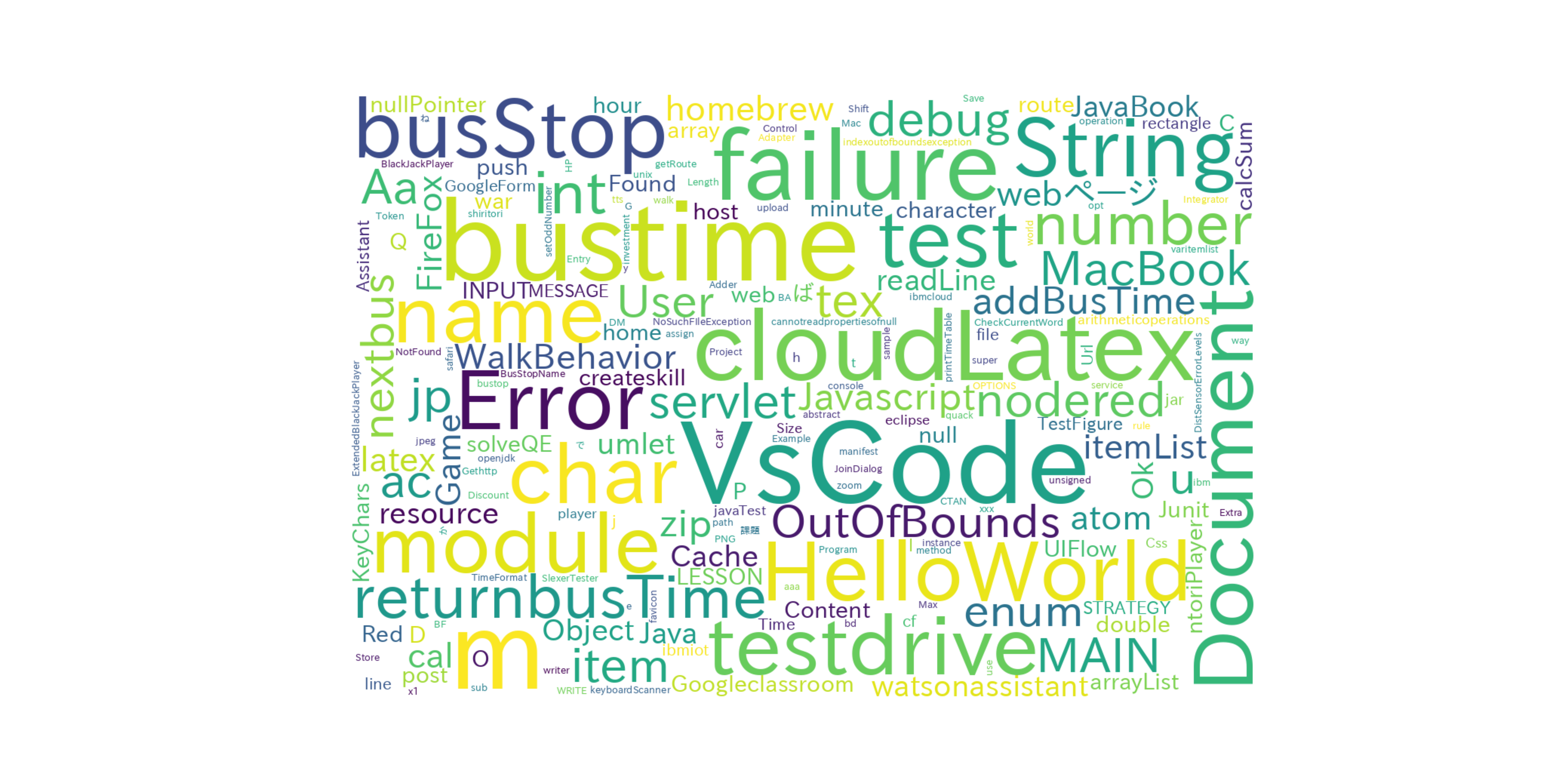}
    \caption{A Word Cloud plot of our collected corpus in the educational computer science domain.}
    \label{word_cloud_edu}
\end{figure}

To better understand the distribution of words in each collected corpus, we visualize the most frequent terms in it using \textit{Word Clouds}. Fig.~\ref{word_cloud_clinic} and \ref{word_cloud_edu} present a Word Cloud representations of the corpus in the clinical domain and the one in the educational computer science domain, respectively. These visualizations provide insight into the key terminologies present in each domain, highlighting domain-specific vocabulary that reflects the primary topics discussed within the corpus.

\subsection{Fine-tuning a Data Generator to Generate Synthetic Sentences}

T5~\cite{raffel2020exploring} is chosen as our data generator which employs a transformer-based encoder-decoder architecture to comprehensively process input sequences and generate coherent and contextually appropriate output sequences. Unlike GPT (Generative Pre-trained Transformer) which adopts a decoder-only model to predict the next word in a sequence~\cite{brown2020language}, T5 uses a text-to-text framework where all tasks are framed as text generation problems. 
T5 is pre-trained on Colossal Clean Crawled Corpus (C4) that is a large-scale dataset of web-sourced general-domain text cleaned to ensure the quality. The training objective of T5 extends masked language modeling by treating the prediction of masked tokens as a sequence generation task that requires appropriate modeling of sequential relations among tokens. 
During training, spans of text (sets of continuous tokens) in an input sequence are randomly selected and each of them is replaced with a single ``sentinel token'' that serves as a placeholder to construct a corrupted input sequence. The corresponding target sequence is formed by replacing unselected spans in the input sequence with sentinel tokens. T5 is trained to predict the target sequence based on the corrupted input sequence. By merging multiple tokens into a single sentinel token, the training process can speed up so that T5 can be trained efficiently on massive C4 dataset. This enables T5 to generate high-quality and contextually rich text. 



Using a specific domain corpus, we fine-tune the pre-trained Japanese T5 Version 1.1 model\footnote{https://huggingface.co/sonoisa/t5-base-japanese-v1.1} which is an improved version of the original T5 model. Referring to the T5 official training setting~\cite{raffel2020exploring}, we randomly mask $15$ percent of tokens in one batch of input sentences and keep the average length of spans as $3$. The cross-entropy loss function is used to measure the accuracy of predicting masked spans with Adafactor optimizer~\cite{shazeer2018adafactor}.

The fine-tuned T5 is used to generate hard negative sentences for contrastive learning on the specific domain. This generation is formulated as the following fill-in-the-blank task: The input is a sentence in the specific domain corpus and multiple spans in it are masked with sentinel tokens. The output is relevant spans to those sentinel tokens. By replacing masked spans with generated ones, we can generate a hard negative sentence that is syntactically similar and semantically dissimilar to the input sentence.
As demonstrated in the preliminary experiment in Section~\ref{ssec:relevant_content_words}, nouns are the most relevant content words in Japanese sentences. Hence, GINZA is used to parse every sentence in the specific domain corpus and POS (Part-Of-Speech) tags are determined by spaCy~\cite{honnibal2015improved}.
Then, we mask ``noun spans'' each of which consists of consecutive noun tokens in order to generate natural and contextually coherent hard negative sentences. 
Figure~\ref{fig1} (b) illustrates that \textless X\textgreater, \textless Y\textgreater \ and \textless Z\textgreater \ in the box annotated with ``Masked noun spans'' are filled with generated spans in the box annotated with ``Fill-in-the-blank''. 

 
Since T5 is primarily trained to reconstruct sentences by predicting masked spans, it is prone to replicating sentences in the specific domain corpus or generating sentences that are similar to them. In other words, generated sentences may be inappropriate as hard negative sentences. To overcome this, our inference adopts a beam search that explores diverse sentences in parallel by maintaining top-$k$ candidate sentences ($k$ is called the beam width).
This approach ensures that generated sentences vary enough from input sentences and can be regarded as hard negative sentences.



\subsection{Contrastive Learning with Hard Negative Sentences}
\label{ssec:our_constrastive_learning}

\begin{figure}[htbp]
    \centering
    \includegraphics[width=\textwidth]{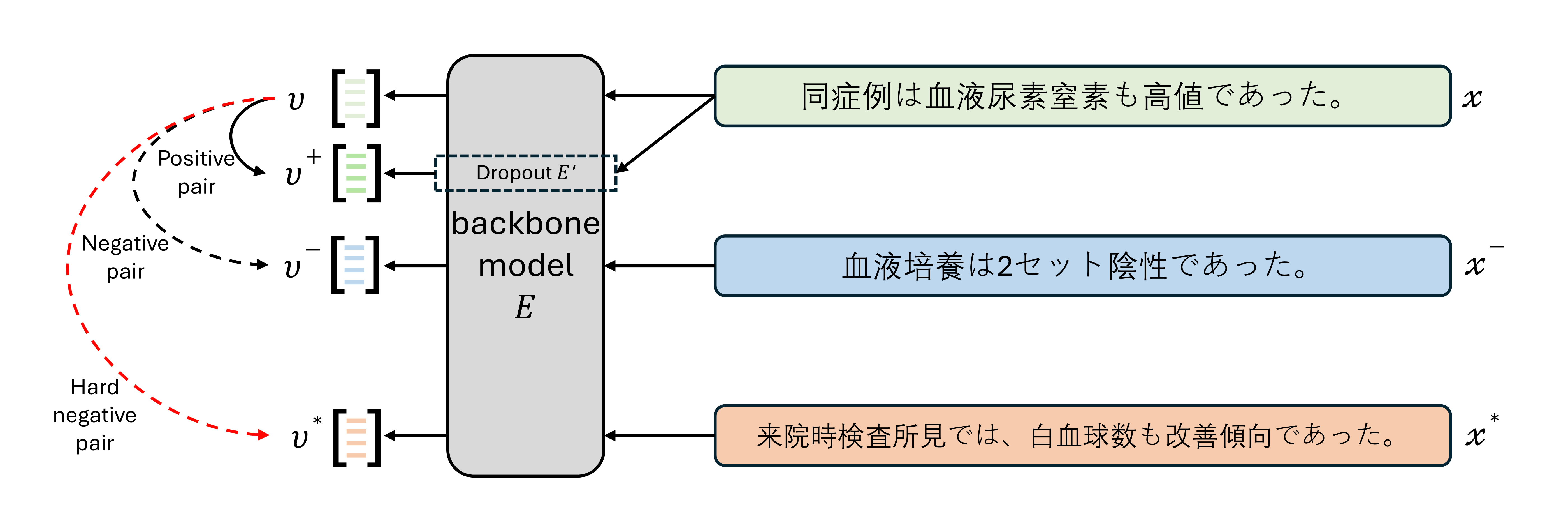}
    \caption{An illustration of the core idea of our contrastive learning with generated hard negative sentences.}
    \label{fig:contrastive_overview}
\end{figure}

By referring to Fig.~\ref{fig:contrastive_overview}, let us define mathematical symbols that are needed to explain our contrastive learning method and describe its core idea for adapting a backbone model. First of all, $x$ denotes an anchor sentence in a specific domain corpus and its embedding $\boldsymbol{v}$ is obtained as $\boldsymbol{v} = E(x)$ meaning that a backbone model $E$ is applied to $x$. Next, inspired by SimCSE~\cite{gao2021simcse}, a \textit{positive embedding} $\boldsymbol{v}^+$ is computed for $x$ as $\boldsymbol{v}^+ = E'(x)$ where $E'$ is a model acquired by performing random unit dropout on $E$. That is, $\boldsymbol{v}$ and $\boldsymbol{v}^+$ are calculated for the same sentence $x$, but they differ in dropout masks which are placed on fully-connected layers as well as attention probabilities as a minimal form of data augmentation using dropout noise~\cite{Transformers} (the dropout rate is set to $0.1$).
This $\boldsymbol{v}^+$ is combined with $\boldsymbol{v}$ to form a positive pair $(\boldsymbol{v}, \boldsymbol{v}^+)$ because both of them are obtained for $x$ and should be similar to each other.

 
Compared to $x$, each of the other sentences in the specific domain corpus is assumed to be semantically dissimilar and regarded as a \textit{negative sentence} $x^-$. Its embedding $\boldsymbol{v}^-$ is calculated using $E$ as $\boldsymbol{v}^- = E(x^-)$. Then, $(\boldsymbol{v}, \boldsymbol{v}^-)$ is regarded as a negative pair so  $\boldsymbol{v}$ and $\boldsymbol{v}^-$ should be different from each other. Moreover, our data generator generates hard negative sentences by replacing noun chunks in $x$. Each of them is represented by $x^*$ and its embedding $\boldsymbol{v}^*$ is obtained by $E$ as $\boldsymbol{v}^* = E(x^*)$. Subsequently, $\boldsymbol{v}^*$ is paired with $\boldsymbol{v}$ as a hard negative pair $(\boldsymbol{v}, \boldsymbol{v}^*)$ since their discrimination is more difficult than the one between $x$ and $x^-$ due to the common syntax of $x$ and $x^*$. Thus, $\boldsymbol{v}^*$ tends to be similar to $\boldsymbol{v}$, but they are required to be different from each other. In other words, our contrastive learning method using the above-mentioned positive, negative and hard negative pairs adapts $E$ so as to produce embeddings that are useful for both coarse-grained and fine-grained discrimination of sentences in the specific domain.

\begin{algorithm}[htbp]
\caption{Contrastive Learning with T5-Generated Hard Negative Sentences}\label{alg:contrastive}
\begin{algorithmic}[1]
\State \textbf{Input:}
\State $E$: Backbone model, T5: Fine-tuned T5 model, $N$: Batch size, $\tau$: Temperature, $\alpha$: Weight for discriminating hard negative pairs
\State \textbf{Output:} \State $E$: Adapted backbone model

\For{each minibatch $X$ with $N$ sentences}
    \State $X^* \gets \text{T5}(X)$ \Comment{Generate hard negative sentences}
    
    \State \textbf{Compute embeddings:}
    \State $V \gets E(X)$ \Comment{$N \times C$ matrix representing $N$ $C$-dimensional embeddings of anchor sentences}
    \State $V^+ \gets E'(X)$ \Comment{$N \times C$ matrix representing $N$ $C$-dimensional positive embeddings based on random unit dropout}
    \State $V^* \gets E(X^*)$ \Comment{$N \times C$ matrix representing $N$ $C$-dimensional embeddings of hard negative sentences}
    
    \State \textbf{Compute the cross entropy loss of Equation (\ref{eq:cross_entropy}):}
    \State $Sim^{+} \gets V V^{+^\top}$ \Comment{$N \times N$ similarity matrix representing pairwise cosine similarities of embeddings in $V$ and $V^+$}
    \State $Sim^{*} \gets V V^{*\top}$ \Comment{$N \times N$ similarity matrix representing pairwise cosine similarities of embeddings in $V$ and $V^*$}
    \State $Sim^{+} \gets Sim^{+} / \tau$, \; $Sim^{*} \gets Sim^{*} / \tau$ \Comment{Incorprate $\tau$ into the similarity matrices} 
    \State $Sim^{*} \gets Sim^{*} + \log(\alpha) \ I$ \Comment{Integrate $\alpha$ in the form of $\log (\alpha)$ with the similarities for hard negative pairs ($I$ is an $N \times N$ identity matrix)}
    \State $Sim \gets \left[ Sim^{+} \| Sim^{*} \right]$ \Comment{$N \times 2N$ similarity matrix obtained by the horizontal concatenation of $Sim^+$ and $Sim^*$}
    \State $\boldsymbol{l} \gets (1, 2, \cdots, N)^\top$ \Comment{$N$-dimensional label vector requiring that the embedding of the $i$th anchor sentence should be similar to the $i$th positive embedding}
    \State $\mathcal{L} \gets \text{CrossEntropyLoss}(Sim, \boldsymbol{l})$

    \State \textbf{Update $E$'s parameters:}
    \State Perform backpropagation based on $\mathcal{L}$ to update $E$'s parameters
\EndFor
\State Return $E$
\end{algorithmic}
\end{algorithm}

Our contrastive learning method is implemented as shown in Algorithm~\ref{alg:contrastive}. Line $5$ presents that a minibatch $X = \{ x_i \}_{i=1}^{N}$ consists of $N$ sentences sampled from the domain specific corpus. Here, $x_i$ is treated as an anchor sentence. Then, as depicted at line $6$, our data generator, fine-tuned T5, is used to generate $X^* = \{ x^*_i \}_{i=1}^{N}$ of $N$ hard negative sentences, each of which $x^*_i$ is the hard negative sentence generated for $x_i$. Subsequently, $N$ anchor sentences in $X$ are fed into a backbone model $E$ to obtain $V = \{ \boldsymbol{v}_i \}_{i=1}^{N}$ consisting of their embeddings (line $8$). Similarly, as described at line $10$, $E$ is used to acquire $V^* = \{ \boldsymbol{v}^*_i \}_{i=1}^{N}$ for $X^*$. On the other hand, a set $V^+ = \{ \boldsymbol{v}_i^+ \}_{i=1}^{N}$ where $\boldsymbol{v}_i^+$ is a positive embedding for $x_i$ and is gained using $E'$ characterized by random unit dropout on $E$. Now, let us consider how to utilize $V$, $V^+$ and $V^*$ in order to efficiently define positive, negative and hard negative pairs for $x_i$. It is clear that $\boldsymbol{v}_i^+$ in $V^+$ and $\boldsymbol{v}_i^*$ in $V^*$ are paired with $\boldsymbol{v}_i$ to form a positive pair $(\boldsymbol{v}_i, \boldsymbol{v}_i^+)$ and a hard negative pair $(\boldsymbol{v}_i, \boldsymbol{v}_i^*)$, respectively. Regarding negative pairs, $x_j$ in $X$ and $x^*_j$ in $X^*$ ($j \neq i$) are regarded as negative sentences for $x_i$. Thus, both $\boldsymbol{v}^+_j$ in $V^+$ and $\boldsymbol{v}_j^*$ in $V^*$ can be used to define negative pairs $(\boldsymbol{v}_i, \boldsymbol{v}_j^+)$ and $(\boldsymbol{v}_i, \boldsymbol{v}_j^*)$. We denote by $P_i^-$ a set of such $2N-2$ negative pairs for $x_i$\footnote{$\boldsymbol{v}_j$ in $V$ is not used to define negative pairs for $x_i$ because of the computational efficiency. That is, a process to exclude $\boldsymbol{v}_i$ is needed to define negative pairs for $x_i$.}. Note that each negative pair in $P_i^-$ is expressed as $(\boldsymbol{v}_i, \boldsymbol{v}_j^-)$ ($1 \leq j \leq 2N -2$) without discriminating between $(\boldsymbol{v}_i, \boldsymbol{v}_j^+)$ and $(\boldsymbol{v}_i, \boldsymbol{v}_j^*)$ in the following discussions.

Using $(\boldsymbol{v}_i, \boldsymbol{v}_i^+)$, $(\boldsymbol{v}_i, \boldsymbol{v}_i^*)$ and $P_i^-$, the cross entropy loss $\mathcal{L}_i$ for $x_i$ is defined as follows:
\begin{equation}
\label{eq:cross_entropy}
\mathcal{L}_i = -\mathrm{log} \ \frac{e^{\mathrm{sim} (\boldsymbol{v}_{i}, \boldsymbol{v}_{i}^+)/\tau } }{ e^{\mathrm{sim} (\boldsymbol{v}_{i}, \boldsymbol{v}_{i}^+)/\tau } + \alpha \ e^{\mathrm{sim} (\boldsymbol{v}_{i}, \boldsymbol{v}_{i}^*)/\tau } + \sum_{ (\boldsymbol{v}_{i}, \boldsymbol{v}_{j}^-) \in P_i } e^{\mathrm{sim} (v_{i},v_{j}^-)/\tau} },
\end{equation}
where $\mathrm{sim} (\cdot, \cdot )$ represents the cosine similarity between two embeddings. In addition, $\tau$ is the temperature which controls the relative distribution of similarities~\cite{wang2021understanding} and $\alpha$ is the weight to control how important the discrimination of $(\boldsymbol{v}_i, \boldsymbol{v}_i^*)$ is. Using $\mathcal{L}_i$, $E$ is adapted so that the similarity for $(\boldsymbol{v}_i, \boldsymbol{v}_i^+)$ is considerably larger than the other similarities for pairs of semantically dissimilar sentences (i.e., $(\boldsymbol{v}_i, \boldsymbol{v}_i^*)$ and $(\boldsymbol{v}_i, \boldsymbol{v}_j^-)$s in $P_i$). Therefore, the accurate discrimination of sentences in the specific domain is achieved.

Lines $11$ to $18$ in Algorithm~\ref{alg:contrastive} shows an efficient batch computation of $\mathcal{L}_i$ for $N$ anchor sentences in a minibatch. The matrix operations at lines $11$ to $14$ produces a $N \times 2N$ matrix $Sim$ where the $i$th row represents $2N$ values corresponding to the exponents of $e$ in Equation (\ref{eq:cross_entropy}). Specifically, except the element $Sim_{i,N+i}$ at the $i$th row and the $(N+i)$th column for the hard negative negative pair, the other elements express cosine similarities divided by $\tau$ and can be directly used as exponents of $e$ in Equation (\ref{eq:cross_entropy}). For the hard negative pair, the cosine similarity divided by $\tau$ is combined with $\log (\alpha)$, that is, $Sim_{i,N+i} = \mathrm{sim} (\boldsymbol{v}_{i}, \boldsymbol{v}_{i}^*)/\tau + \log(\alpha)$. The term $\alpha \ e^{\mathrm{sim} (\boldsymbol{v}_{i}, \boldsymbol{v}_{i}^*)/\tau }$ in Equation (\ref{eq:cross_entropy}) is obtained using $Sim_{i,N+i}$ as an exponent of $e$, that is, $e^{\mathrm{sim} (\boldsymbol{v}_{i}, \boldsymbol{v}_{i}^*)/\tau + \log(\alpha)} = e^{\mathrm{sim} (\boldsymbol{v}_{i}, \boldsymbol{v}_{i}^*)/\tau} \cdot e^{\log(\alpha)} = \alpha \ e^{\mathrm{sim} (\boldsymbol{v}_{i}, \boldsymbol{v}_{i}^*)/\tau}$. Finally, the losses for $N$ anchor sentences are computed by passing $Sim$ to a cross entropy loss function implemented in a standard deep learning library like PyTorch. Here, a label vector $\boldsymbol{l}$ containing integers increasing from $1$ is also given to the loss function, so that the element for each positive pair in $Sim$ is required to be larger than the elements for the other pairs. 

\section{Japanese STS Benchmark}
\label{sec:jsts_benchmark}

This section presents our Japanese STS (JSTS) benchmark study by firstly describing the construction of our comprehensive JSTS benchmark dataset, and then showing results to evaluate various existing sentence representation methods and backbone models on it.

\subsection{Japanese STS Benchmark Dataset}


Our comprehensive JSTS benchmark dataset is constructed by combining  JSICK~\cite{yanaka2022compositional} and JSTS~\cite{kurihara2022jglue} datasets with translated STS12-16 and STS-B datasets using machine translation. The translated datasets are derived from original English datasets. We apply a similar translation strategy by referring to \cite{yoshikoshi2020multilingualization} which uses machine translation followed by filtering to balance the quality and scale of a dataset effectively.


First, we translate each of STS12-16 and STS-B datasets into Japanese using Google translate API\footnote{https://cloud.google.com/translate/} and DeepL translate API\footnote{https://www.deepl.com/en/pro-api}. Then, the translated sentences are automatically filtered by comparing each of the original English sentences to the sentence that is back-translated from the corresponding translated Japanese sentence. If these two sentences are similar, the translation is regarded as adequate, otherwise it is inadequate and the translated Japanese sentence is filtered out. To implement this filtering, the back-translation is performed by the same machine translator used for the English-Japanese translation, and the sentence comparison is based on BLEU score~\cite{papineni2002bleu} that examines how many N-grams are matching between an original sentence and the back-translated sentence. With respect to this, we find that most of BLEU scores are $0$ when $N>1$, so we just apply BLEU1 scores which consider individual word matches.


To evaluate the quality of a translated dataset and decide which of Google or DeepL translators is used, by referring to \cite{yoshikoshi2020multilingualization}, we conduct a relative comparison among translated datasets in terms of linear regression performances. Each translated dataset has a different quality and size that influence the regression performance. We take advantage of this as follows: If the performance of a linear regression model trained on one translated dataset is better than the performances of linear regression models trained on the other translated datasets, the former can be considered more consistent than the latter. More closely, using a dataset containing translated sentences defined by a different BLUE1 score threshold, we build a linear regression model that consists of a pre-trained BERT model to extract CLS token's embedding for a sentence pair and a linear regression layer to predict the sentence similarity score for the sentence pair. Both of the BERT model and linear regression layer are optimized using the translated dataset. Here, the same training setting is used to train linear regression models for all the translated datasets (i.e., the epoch number is $3$ and the learning rate is $5e-5$). Then, a trained model is tested on JSTS dataset because it is manually created and can be seen as gold labels.

The results are shown in Table~\ref{tab1} where Mean Absolute Error (MAE), Mean Squared Error (MSE) and Coefficient of Determination (R-Squared (R2)) are used as the evaluation metrics. MAE and MSE measure the average of the absolute differences between predicted and ground truth values and the squared differences, respectively.
R2 is a statistical measure that represents the squared correlation coefficient between ground truth values and embeddings obtained by applying the pre-trained BERT to sentences in a translated dataset. We chose the dataset that is translated by DeepL and the no BLUE1 threshold because it leads to the best performance (i.e., the lowest MAE and MSE, and the highest R2) as shown in Table~\ref{tab1}. 
Here, translated sentences whose BLEU1 score is $0$ are filtered out. We check these filtered sentences and discover issues such as gibberish and ambiguity. Afterward, we manually discard some untranslated sentences for unknown reasons to get the last edition of the translated STS datasets. The performance on this last edition is shown at the bottom of Table \ref{tab1} and we believe it is valid compared to the other machine-translated datasets. Finally, Table~\ref{tab2} shows the statistics of the last edition of our comprehensive JSTS benchmark dataset where the translated datasets chosen above are combined with human-annotated published JSICK~\cite{yanaka2022compositional} and JSTS~\cite{kurihara2022jglue} datasets.

\begin{table}[htbp]
\centering
\begin{adjustbox}{width=\textwidth}
\begin{tabular}{llccccccc}
\hline \hline
\multicolumn{2}{l}{\textbf{Google translator}}  &        &        &        &        &        &        &        \\ \hline
\multicolumn{2}{l|}{BLEU   threshold}  & -      & 0.05   & 0.1    & 0.15   & 0.2    & 0.25   & 0.3    \\ \hline
datasize     & \multicolumn{1}{l|}{}   & 21172  & 20612  & 20449  & 19508  & 18207  & 16694  & 14904  \\ \hline
MAE          & \multicolumn{1}{l|}{}   & 0.4415 & 0.4443 & 0.4807 & 0.4493 & 0.4963 & 0.4966 & 0.4916 \\ \hline
MSE          & \multicolumn{1}{l|}{}   & 0.3384 & 0.336  & 0.3793 & 0.3332 & 0.412  & 0.4191 & 0.4018 \\ \hline
R2           & \multicolumn{1}{l|}{}   & 0.6481 & 0.6505 & 0.6058 & 0.6537 & 0.5727 & 0.5632 & 0.5822 \\ \hline\hline
\multicolumn{2}{l}{\textbf{DeepL   translator}} &        &        &        &        &        &        &        \\ \hline
\multicolumn{2}{l|}{BLEU   threshold}  & -      & 0.05   & 0.1    & 0.15   & 0.2    & 0.25   & 0.3    \\ \hline
datasize     & \multicolumn{1}{l|}{}   & \textbf{21172}  & 19337  & 19144  & 18368  & 17535  & 16683  & 15727  \\ \hline
MAE          & \multicolumn{1}{l|}{}   &\textbf{0.4206} & 0.433  & 0.4314 & 0.4397 & 0.451  & 0.4448 & 0.442  \\ \hline
MSE          & \multicolumn{1}{l|}{}   &\textbf{0.3058} & 0.3135 & 0.3117 & 0.324  & 0.342  & 0.3312 & 0.328  \\ \hline
R2           & \multicolumn{1}{l|}{}   & \textbf{0.6894} & 0.6802 & 0.6834 & 0.6711 & 0.6529 & 0.6638 & 0.6666 \\ \hline\hline
\multicolumn{2}{l}{\textbf{Last edition}} &        &        &        &        &        &        &        \\ \hline
datasize     &       &  19330 & MAE    & \textbf{0.4119} & MSE    & \textbf{0.2821} & R2     & \textbf{0.7137} \\ \hline\hline
\end{tabular}
\end{adjustbox}
\caption{The performances of linear regression models trained on different translation data. Mean Absolute Error (MAE), Mean Square Error (MSE) and R-Squared (R2) are used as metrics to evaluate linear regression performances that reflect translation qualities.}
\label{tab1}
\end{table}


\begin{table}[htbp]
\centering
\begin{tabular}{l|c|c}
\hline
Japanese   STS benchmark          & Translated by & \#Examples \\ \hline
\textbf{Total}                    & -             & 43165      \\
\textbf{STS12}                    & Machine       & 2615       \\
\textbf{STS13}                    & Machine       & 1063       \\
\textbf{STS14}                    & Machine       & 3165       \\
\textbf{STS15}                    & Machine       & 2901       \\
\textbf{STS16}                    & Machine       & 1143       \\
\textbf{STS-B}                    & Machine       & 8443       \\
\textbf{JSICK}                    & Human         & 9927       \\
\textbf{JSTS}                     & Human         & 13908      \\ \hline
\end{tabular}
\caption{Statistics of our comprehensive JSTS benchmark dataset.}
\label{tab2}
\end{table}

\subsection{Benchmark Results}
\label{ssec:j_benchmark_result}

We compare existing sentence representation methods and backbone models pre-trained on the general domain using our comprehensive JSTS benchmark dataset. To make the following discussions clear, we call a combination of a sentence representation method and a backbone model a \textit{sentence embedding model}. Because there are no pre-trained sentence embedding models specialized to Japanese and there are several unknown implementation details of public checkpoints, we train them and examine their performances by ourselves. The sentence embedding models in this experiment adopt the following sentence representation methods:



\begin{description}
    \item[FastText~\cite{joulin2017fasttext}:] FastText is a word-view method that generates word embeddings using subword information to handle out-of-vocabulary words effectively while maintaining computational efficiency. The embedding of a sentence is computed by averaging the embeddings of words in it.
    \item[Average BERT:] Average BERT is a word-view method that creates a sentence embedding by averaging the contextualized token embeddings from a pre-trained BERT model~\cite{devlin2018bert}, providing a simple, limited representation of a sentence.
    \item[BERT-flow~\cite{li2020sentence}:] BERT-flow is a word-view method that performs flow-based linear transformation as post-processing, where a sentence embedding is computed by transforming the average of token embeddings at the last two layers of a pre-trained BERT into an isotropic space.
    \item[BERT-whitening~\cite{su2021whitening}:] BERT-whitening is a word-view method using a whitening transformation that normalizes the distribution of sentence embeddings obtained by averaging token embeddings at the last two layers of a pre-trained BERT. 
    \item[Sentence-BERT~\cite{reimers-gurevych-2019-sentence}:] Sentence-BERT is a supervised sentence representation method that fine-tunes a pre-trained BERT model using a Siamese architecture to generate sentence embeddings, which are explicitly optimized for STS tasks. Then, using this BERT, the embedding of a sentence is obtained by average pooling on last layer's outputs for all tokens in the sentence.
    \item[SimCSE~\cite{gao2021simcse}:] SimCSE is employs contrastive learning with dropout-based augmentation to produce high-quality sentence embeddings in an unsupervised or supervised framework.
    \item[MixCSE~\cite{zhang2022unsupervised}:] MixCSE is an unsupervised contrastive learning method that artificially constructs hard negative embeddings by mixing both positive and random negative embeddings. 
    \item[DiffCSE~\cite{chuang2022diffcse}:] DiffCSE is an unsupervised method that combines contrastive learning with a difference prediction objective in order to train sentence embeddings that are sensitive to meaningful edits by leveraging masked language modeling and a conditional discriminator.
\end{description}


For each method, an official code is used to train a sentence embedding model. We tested various hyperparameters and find that the officially recommended hyperparameters are preferable to others in most situations. Thus, all the above-mentioned methods are run on the officially recommended settings.
In addition, all the methods train BERT-base or BERT-large as a backbone model using their public checkpoints\footnote{https://huggingface.co/tohoku-nlp}.


For unsupervised contrastive learning methods, the following three types of Wikipedia datasets are used for training. The first one is \textit{wiki1m} that is collected by ourselves from the official Japanese Wikipedia corpus\footnote{https://dumps.wikimedia.org/jawiki/} using WikiExtractor\footnote{https://github.com/attardi/wikiextractor} and GINZA. The second and third types of datasets, \textit{wiki40b}\footnote{https://huggingface.co/datasets/google/wiki40b} and \textit{wikipedia}\footnote{https://huggingface.co/datasets/wikimedia/wikipedia} are publicly available on Huggingface. For all of these Wikipedia datasets, $10^{6}$ sentences are randomly sampled as training data. Supervised methods use JSNLI dataset that contains $533,005$ sentence pairs after filtering. For BERT-flow and BERT-whitening we use JSNLI dataset to train transformation models. 



\begin{sidewaystable}
  \begin{tabular}{c|l|cccccc|cc|c}
\hline
& Sentence embedding model & \multicolumn{6}{c|}{Machine translation} & \multicolumn{2}{|c|}{Human-annotated} & \\
Type & $<$Method$>$-$<$Backbone$>$              & \textbf{STS12} & \textbf{STS13} & \textbf{STS14} & \textbf{STS15} & \textbf{STS16} & \textbf{STS-B} & \textbf{JSICK} & \textbf{JSTS} & \textbf{Avg.} \\ \hline
\multirow{8}{*}{\shortstack[c]{Word-view}} & FastText  & 42.81          & 51.62          & 42.34          & 54.22          & 54.02          & 51.27          & 74.76          & 60.17        & 53.90         \\
& Average BERT-base &40.04 &50.86 &36.23 &51.61 &54.15 &51.39 &71.97 &65.19 &52.68 \\ 
& Average BERT-base-v2 &41.7 &50.34 &36.84 &51.51 &53.6 &50.88 &74.33 &66.42 &53.2 \\
& Average BERT-base-v3 &40.93 &50.25 &37.27 &51.75 &54.73 &54.28 &75.28 &69.4 &54.24 \\
& Average BERT-large &39.69 &48.33 &35.42 &49.47 &53.26 &45.74 &65.33 &65.98 &50.4 \\
& Average BERT-large-v2 &39.48 &49.76 &35.5 &50.37 &53.8 &50.58 &74.76 &66.7 &52.62 \\
& BERT-flow-base       & 41.83          & 52.32          & 39.16          & 53.07          & 55.01          & 55.09          & 77.26          & 64.90         & 54.83         \\
& BERT-whitening-base         & 38.62          & 51.72          & 37.69          & 55.05          & 56.94          & 46.26          & 71.86          & 56.00         & 51.77         \\ \hline
\multirow{11}{*}{\shortstack[c]{Unsupervised\\contrastive\\learning}} & SimCSE-BERT-base-v3 (\textit{wiki1m})            & 52.1          & 58.84          & 48.92   &64.31          & 63.01          & 70.55          & 77.97          & 75.1         & 63.85         \\
& SimCSE-BERT-large-v2 (\textit{wiki1m})           & \textbf{52.24}          & 59.66          & \textbf{51.44}        & \textbf{67.93}         & \textbf{64.43}          & 70.99          & 78.51          & 75.04     & \textbf{65.03} \\
& SimCSE-BERT-base-v3 (\textit{wiki40b}) &48.63 &57.37 &49.08 &64.24 &61.34 &70.44 &76.24 &73.32 &62.58 \\
& SimCSE-BERT-large-v2 (\textit{wiki40b}) &50.14 &56.13 &49.51 &65.27 &61.35 &69.82 &79.68 &75.74 &63.46 \\
& SimCSE-BERT-base-v3 (\textit{wikipedia}) &51.85 &57.71 &49.29 &61.92 &60.06 &71.14 &75.1 &74.49 &62.69 \\
& SimCSE-BERT-large-v2 (\textit{wikipedia}) &52.09 &56.93 &50.43 &65.5 &61.31 &71.43 &78.14 &\textbf{76.83} &64.08 \\
& MixCSE-BERT-base (\textit{wiki1m})  &51.23 &58.24 &48.42 &62.5 &61.44 &70.21 &76.97 &71.85 &62.61 \\
& MixCSE-BERT-base-v3 (\textit{wiki1m}) &52.2 & \textbf{59.67} &50.13 &66.51 &63.79 &71.63 &77.73 &73.98 &64.45 \\
& MixCSE-BERT-large (\textit{wiki1m})    &50.22 &56.89 &48.7 &62.87 &61.18 &67.76 &78.15 &73.89 &62.46 \\
& MixCSE-BERT-large-v2 (\textit{wiki1m}) &49.93 &58.45 &51 &65.45 &62.01 &\textbf{72.92} &\textbf{80.77} &76.47 &64.62 \\
& DiffCSE-BERT-base (\textit{wiki1m})   &50.09 &56.15 &46.14 &61.87 &60.44 &69.37 &76.29 &71.07 &61.43        \\ \hline
\multirow{2}{*}{\shortstack[c]{Supervised}} & Sentence-BERT-base   &51.8 &52.92 &48.86 &60.48 &\textbf{58.53} &64.76 &65.25 &75.28 &59.73 \\
& Sentence-BERT-large  &52.56 & \textbf{54.06} & \textbf{49.4} &\textbf{62.45} &58.46 &\textbf{66.17} &67.68 &\textbf{76.84} &\textbf{60.95} \\ \hline
\multirow{5}{*}{\shortstack[c]{Supervised\\contrastive\\learning}} & SimCSE-BERT-base    &48.04 &50.17 &39.4 &51.05 &52.61 &57.96 &75.01 &66.91 &55.14\\
& SimCSE-BERT-base-v2 &52.19 &52.69 &44.02 &55.12 &57.27 &59.76 &77.84 &69.36 &58.53 \\
& SimCSE-BERT-base-v3 &52.13 &53.79 &46.41 &57.39 &58.42 &61.91 &\textbf{78.49} &72.79 &60.17 \\
& SimCSE-BERT-large   &47.5 &49.5 &43.7 &50.12 &52.41 &56.28 &76.57 &69.12 &55.65     \\ 
& SimCSE-BERT-large-v2 & \textbf{52.59} &50.79 &44.68 &57.4 &55.25 &59.45 &76.6 &74.81 &58.95 \\
\hline
  \end{tabular}
\caption{Performances of different sentence embedding models on our comprehensive JSTS benchmark dataset.}
\label{tab3}
\end{sidewaystable}


Table~\ref{tab3} shows the evaluation results of different sentence embedding models defined by combining the sentence representation methods and pre-trained backbone models described above. To measure their performances, we apply Spearman's rank correlation as the metric which shows the correlation between the ranking of predicted similarity scores and the one of human-annotated scores for sentence pairs. By referring to previous work~\cite{gao2021simcse, zhang2022unsupervised, chuang2022diffcse}, we address ``all'' setting where Spearman's rank correlation is computed over all sentence pairs in a dataset, rather than separately computing it over sentence pairs in individual subsets.


First, the sentence embedding models based on unsupervised contrastive learning in the middle section of Table~\ref{tab3} (i.e., SimCSE-BERT-base-v3 to DiffCSE-BERT-base) perform better than the ones based on word-view methods. As highlighted in bold, the former show the best or nearly the best performances on almost all STS tasks. Based on English STS benchmark experiments~\cite{gao2021simcse}, the sentence embedding models based on supervised contrastive learning in the five rows from the bottom of Table~\ref{tab3} are expected to outperform Sentence-BERT models, but the latter outperform the former. In addition, the sentence embedding models based on supervised contrastive learning are expected to outperform the ones based on unsupervised contrastive learning, but this is not the case in our experiment. The reason is that a much larger amount of labeled sentence pairs are available in English experiments compared to JSNLI dataset. Specifically, JSNLI dataset is just about half of training data consisting of SNLI \cite{bowman2015large} and MNLI \cite{williams2018broad} datasets in the English experiments \cite{gao2021simcse}. This point was overlooked in other research but uncovered in our experiment. Despite this observation, the sentence embedding models based on supervised contrastive learning get nearly the best performances even though labeled training data are limited.



Overall, both of supervised and unsupervised contrastive learning methods work well for Japanese sentence embedding. To gain the benefit of supervised contrastive learning methods, we generate hard negative sentences to increase labeled sentence pairs.



In addition, we get the following significant indication about the quality of our machine-translated datasets: After checking Table \ref{tab3}, we can find that the results on the human-annotated datasets and our machine-translated datasets indicate the same trend, meaning that sentence embedding models which perform well on the human-annotated datasets also perform well on our machine-translated ones.
This consistent trend reflects our machine-translated datasets serve as a reliable benchmark, comparable to human-annotated datasets. In other words, the semantic relationships captured in our machine-translated datasets are as effective in evaluating model performances as those in human-annotated datasets.

Finally, the following two observations are useful for devising SDJC: The latest pre-trained checkpoints (i.e., \textit{BERT-base-v3} and \textit{BERT-large-v2}) are definitely better than previous ones. In addition, referring to the multiple Wikipedia datasets collected above, the performance difference trained on them is tiny, so we used our own \textit{wiki1m} dataset at the experiments in the next section.


\section{Experiments}
\label{sec4}

This section presents the experiments to devise and evaluate SDJC. Section~\ref{ssec:downstream_tasks} describes the two downstream tasks addressed in this paper. The experiment in Section~\ref{ssec:relevant_content_words} is dedicated to deciding which words in a sentence should be replaced by a data generator to generate hard negative sentences. Implementation details of SDJC are then explained in Section~\ref{ssec:implementation_details}. Afterwards, SDJC's domain adaptation is evaluated through its performance comparison to various related methods in Section~\ref{ssec:main_results}. Finally, Section~\ref{ssec:visualization_se} visually demonstrates the effectiveness of embeddings that are computed by a backbone model adapted by SDJC.

\subsection{Domain-specific Downstream Tasks}
\label{ssec:downstream_tasks}

SDJC is evaluated on two Japanese domain-specific downstream tasks, STS on JACSTS dataset~\cite{mutinda2021semantic} in a clinical domain and information retrieval on QABot dataset~\cite{chen2022retrieval,ide2023qabot} in an educational domain. These tasks are detailed below:

\subsubsection{STS in a Clinical Domain}

\begin{figure}[htbp]
    \centering
    \includegraphics[width=0.5\textwidth]{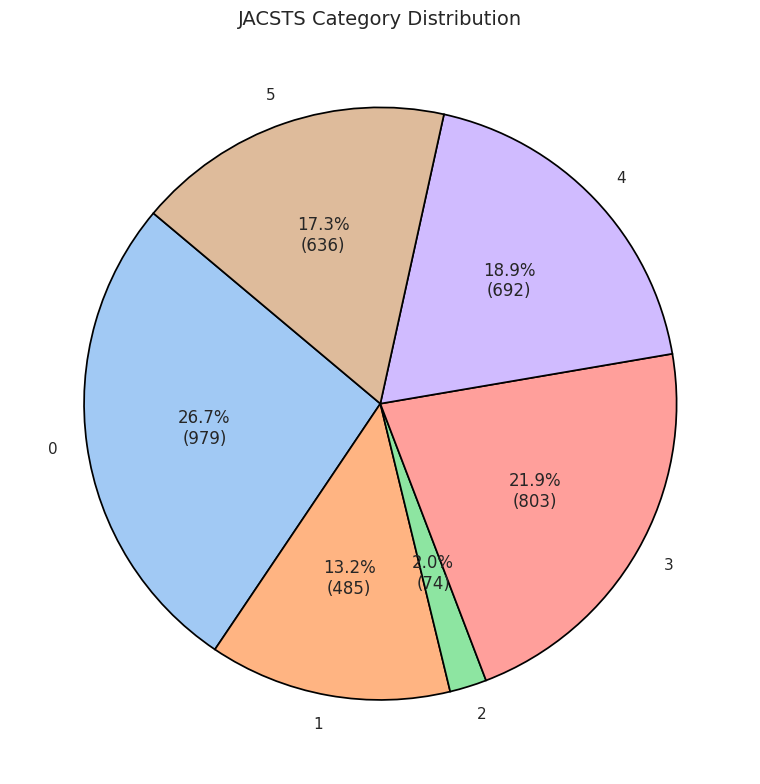}
    \caption{A pie chart illustration of the distribution of JACSTS dataset. Categories corresponding to labeled similarity scores are represented by differently colored segments, with the percentages and sentence pair counts annotated on the pie chart.}
    \label{fig4}
\end{figure}

JACSTS dataset~\cite{mutinda2021semantic} is a publicly available dataset\footnote{https://github.com/sociocom/Japanese-Clinical-STS}  for sentence-level clinical STS from Japanese case reports. The dataset consists of 3,670 sentence pairs, each annotated with a semantic similarity score from $0$ to $5$, where $0$ and $5$ indicate that two sentences are semantically very dissimilar and very similar, respectively. Table \ref{tab_JACSTS} display some examples of sentence pairs with human-annotated scores. Fig.~\ref{fig4} shows the statistical information of JACSTS dataset through a pie chart representation. Here, $26.7$\% sentence pairs are labeled as $0$ and only $2$\% sentence pairs are labeled as $2$.
JACSTS dataset is combined into SentEval toolkit~\cite{conneau2018senteval} where SDJC adapts a backbone model to predict the cosine similarity between embeddings of a sentence pair. The evaluation is performed using Spearman's rank correlation that measures the correlation between the ranking of predicted similarity scores and and the one of human-annotated scores.
Using Spearman's rank correlation, we can assess how well sentence embeddings represented by a adapted backbone model lead to similarity scores that align with human-annotated scores. 


\begin{CJK}{UTF8}{min} 
\begin{table}[htbp]
\centering
\begin{tabular}{p{0.4\textwidth}|p{0.4\textwidth}|c}
\textbf{Sentence 1} & \textbf{Sentence 2} & \textbf{Score}\\
\midrule
しかし股関節周囲筋の多くが切離されていることから、股関節の安定性が望めない &
しかし伸展に関しては痙痛が出現していることから、負荷量や運動範囲を考慮する必要がある & \multirow{2}{*}{\shortstack[c]{0}}\\
However, since much of the periprosthetic hip muscle is transected, hip stability is not desired. &
However, since spastic pain has appeared with respect to extension, it is necessary to consider the amount of load and range of motion. &  \\
\midrule
家族歴:のように, 家系内に難聴, 耳痩孔, 腎疾患を有するものが多い &
家族歴:側頸痩, 耳痩孔, 難聴, 腎病変をはじあとして奇形を有するものはない & \multirow{2}{*}{\centering 1} \\
Family history: many have hearing loss, otosclerosis, and renal disease in the family, as in the following & 
Family history: no malformations, including lateral cervical dystrophy, auriculopathy, hearing loss, or renal involvement & \\
\midrule
歯肉出血を主訴に来院したが, 高血圧症のため4年前よりニフェジピンを服用しており高度な歯肉増殖を認めた &
臨床診断:4年間ニフェジピンを服用しており, 歯肉増殖が著しいことからニフェジピン歯肉増殖症を伴う慢性歯周炎と診断した & \multirow{2}{*}{\centering 2} \\
A patient came to our hospital with a chief complaint of gingival bleeding. He had been taking nifedipine for 4 years for hypertension and showed severe gingival hyperplasia. &
Clinical diagnosis: Chronic periodontitis with nifedipine gingival hyperplasia was diagnosed because the patient had been taking nifedipine for 4 years and had marked gingival hyperplasia. & \\
\midrule
左耳下部に45mm大, 弾性硬, 可動性良好な腫瘤を触知した &
局所所見:左耳下部に45×33mm, 弾性硬な腫瘤を触知した & \multirow{2}{*}{\centering 3} \\
A 45 mm large, elastic, hard, mobile mass was palpated in the lower left ear. &
Local findings: palpable 45 x 33 mm, elastic and firm mass in the lower left ear. & \\
\midrule
手術時間は4時間33分, 麻酔時間は5時間32分であった &
手術時間は3時間30分, 麻酔時間は4時間37分であった & \multirow{2}{*}{\centering 4} \\
The operation time was 4 hours and 33 minutes, and the anesthesia time was 5 hours and 32 minutes. & The operation time was 3 hours and 30 minutes, and the anesthesia time was 4 hours and 37 minutes. & \\
\midrule
腫瘤の頭側および尾側に索状物が付着し, 迷走神経に連続していた &
この腫瘤の頭側および尾側に索状のものが付着し, 迷走神経に連なっていた &
\multirow{2}{*}{\centering 5} \\
The Cordate is attached to the cephalic and caudal sides of the mass and continues with the vagus nerve. &
Cord-like attachments were found on the cephalic and caudal sides of this mass, which were connected to the vagus nerve. & \\
\bottomrule
\end{tabular}
\caption{Examples of sentence pairs in JACSTS dataset. Each score ranging from $0$ to $5$ represents a semantic similarity score between two sentences. Here, $0$ and $5$ indicate that two sentences are very dissimilar and highly similar, respectively.}
\label{tab_JACSTS}
\end{table}
\end{CJK}

\subsubsection{Information Retrieval in an Educational Domain}


Targeting programming exercise courses at Kindai University in Japan, we aim to develop an information retrieval system that searches over a dataset of question-answer (QA) pairs collected by QABot~\cite{ide2023qabot} described in Section~\ref{ssec:specific_domain_corpus_collection}~\cite{chen2022retrieval}. We call this dataset \textit{QABot dataset}. Our system enables a student to find relevant QA pairs each of which consists of a question similar to his/her query and an answer (solution) to it. However, searching over large-scale QABot dataset is time-consuming and tedious, and it often happens to fail finding relevant QA pairs despite a long-time search process. Addressing this point, our system adopts an approach that computes a sentence embedding of the question of each QA pair and retrieves relevant QA pairs to a query in terms of their embeddings. QABot dataset consists of $1,142$ QA pairs that are individually annotated with relevance to each of $20$ queries. As exemplified in Table~\ref{tab6}, for each query, there is at least one question labeled as $1$, indicating that this question is relevant to the query. On the other hand, questions labeled as $0$ are irrelevant to the question. The average number of relevant questions over $20$ queries is $2.65$.




SDJC firstly adapts a backbone model to the Japanese educational computer science domain using the corpus collected in Section~\ref{ssec:specific_domain_corpus_collection}. Then, the adapted backbone model is incorporated into a two-tower architecture by referring to previous studies~\cite{ma2021zero, liang2020embedding, karpukhin2020dense}. The first and second towers share the adapted backbone model and compute the embedding of a query and the one of a question, respectively. Finally, retrieval is performed with respect to cosine similarities between embeddings of questions and that of the query. The performance is evaluated using a Mean Average Precision (MAP), Mean Reciprocal Rank (MRR) and Precision at N (P@N). The MAP represents the mean of average precision scores across all queries to measure the overall ranking quality by considering both precision and recall. The MRR indicates the average of reciprocal ranks of the first relevant question to each of $20$ queries, and represents how quickly relevant questions are retrieved. P@N implies the proportion of relevant questions found in the top $N$ retrieved ones to reflect the precision at a specific cutoff point. These evaluation measures express whether relevant questions to each query are ranked at high positions in a retrieval result which is influenced by the quality of sentence embeddings.


\begin{CJK}{UTF8}{min} 
\begin{table}[htbp]
\centering
\begin{adjustbox}{width=\textwidth}
\begin{tabular}{p{0.9\textwidth} c} 
\makecell[l]{\textbf{Query: 課題1を提出したんですが確認するとtoo largeと出ますなぜですか？} \\ \textit{Query: I submitted assignment 1, but when I check it, it says TOO LARGE.}} \\
\midrule
\textbf{Question} & \textbf{Label} \\
\midrule
課題１を提出して確認したところ，メッセージの最後に”File too large”と出ました．これは提出が失敗していますか？ & \multirow{2}{*}{\large 1} \\
\textit{\footnotesize I submitted and checked Assignment 1, and at the end of the message I got “File too large”. Is this a submission failure?} & \\
課題01-1を提出して確認したのですが,File too largeとなってしまいました.どう対処すれば良いのでしょうか & \multirow{2}{*}{\large 1} \\
\textit{\footnotesize I have submitted and checked Assignment 01-1, but the file is too large. What should I do?} & \\
課題01-1が終わり、実行結果も緑でOKでしたが、ファイルを提出して採点結果確認をしたところ exam1/HelloPM2Test.java hello/HelloPM2.java File too large となり、OKもfailuresも表示されませんでした。 jarファイルの中身を確認したところ、 META-INF/MANIFEST.MF exam1/HelloPM2Test.class exam1/HelloPM2Test.java hello/HelloPM2.class hello/HelloPM2.java の5つのファイルが入っていましたが、正しいですか？ & \multirow{2}{*}{\large 1} \\
\textit{\footnotesize I finished assignment 01-1 and the run result was green and OK, but when I submitted the file and checked the grading results, exam1/HelloPM2Test.java hello/HelloPM2.java File too large was displayed and neither OK nor failures were shown. I checked the contents of the jar file and found 5 files, META-INF/MANIFEST.MF exam1/HelloPM2Test.class exam1/HelloPM2Test.java hello/HelloPM2.class hello/HelloPM2.java. Is this correct?} & \\
課題1の採点結果がまだFile too largeとなるのですが，何が原因なのでしょうか？ & \multirow{2}{*}{\large 1} \\
\textit{\footnotesize I am still getting File too large results for assignment 1.} & \\
\midrule
課題1のTestDriveで実行したいんですが実行しようとするとJUitテストと出てくるんですがどうすればいいですか？ & \multirow{2}{*}{\large 0} \\
\textit{\footnotesize I want to run it on TestDrive for issue 1, but when I try to run it, it comes up as a JUit test, what should I do?} & \\
課題１のテストが成功しているのにも関わらず, ホームページに課題1のjarファイルを提出したらtest.failedと出てしまうのですがなぜでしょうか. jarファイルの中身は確認して中身があることは確認できています & \multirow{2}{*}{\large 0} \\
\textit{\footnotesize Why do I get the error message “test.failed” when I submit the jar file of assignment 1 to the homepage, even though the test of assignment 1 was successful? I have checked the contents of the jar file and confirmed that the contents are there.} & \\
TempHumidWarning.javaのupdateメソッドの引数のpressureとはなんですか？& \multirow{2}{*}{\large 0} \\
\textit{\footnotesize What is the pressure argument of the update method in TempHumidWarning.java?} & \\
課題11-3のコンストラクタのところで、 lines の要素数を天体数として numOfStars に代入 をしたいのですが、numOfStars =lines[numOfStars];にするとエラーが出でくるのですが、どのようにしたらいいですか。& \multirow{2}{*}{\large 0} \\
\textit{\footnotesize In the constructor of issue 11-3, I want to assign the number of elements of lines to numOfStars as the number of celestial bodies, but I get an error when I set numOfStars =lines[numOfStars];.} & \\
\bottomrule
\end{tabular}
\end{adjustbox}
\caption{An example showing a query and relevant and irrelevant questions of QA pairs in QABot dataset. Labels $1$ and $0$ indicate that a question is relevant and irrelevant to the query, respectively.}
\label{tab6}
\end{table}
\end{CJK}

\subsection{Relevant Content Words}
\label{ssec:relevant_content_words}


This experiment aims to identify the POS tag of relevant content words in Japanese. As described in Section~\ref{ssec:contrastive_learning}, relevant content words have big influences on the semantic similarity of  a sentence pair. By replacing relevant content words characterized by the identified POS tag in an anchor sentence, our data generator can safely generate a hard negative sentence that has the same syntactic structure to the anchor sentence but conveys a dissimilar semantic meaning.

To identify the POS tag of relevant content words, we borrow the approach used in \cite{wang2021tsdae}. Specifically, semantically similar sentence pairs are firstly collected, and then the following equation is used to find the word $\hat{w}$ that maximally reduces the cosine similarity $\mathrm{sim}(a,b)$ between semantically similar sentences $a$ and $b$:
\begin{equation}\label{eq5}
    \hat{w} = \mathrm{argmax} _{w} \left( \mathrm{sim}(a,b)-\mathrm{min} \left( \mathrm{sim}(a\setminus w,b),\mathrm{sim}(a,b\setminus w) \right) \right)
\end{equation}
where $w$ is a word that appears in either $a$ or $b$. In Eq.~\ref{eq5}, $\mathrm{sim}(a\setminus w,b)$ represents the cosine similarity between the sentence obtained by deleting $w$ from $a$, and $b$. The same logic applies to $\mathrm{sim}(a,b\setminus w)$. It should be noted that if $w$ appears only in $a$ (or $b$), accordingly only $\mathrm{sim}(a\setminus w,b)$ (or $\mathrm{sim}(a,b\setminus w)$) is counted in Eq.~\ref{eq5}. Finally, we detect $\hat{w}$'s POS tag using spaCy~\cite{honnibal2015improved} and compute the distribution of $\hat{w}$'s POS tags for all sentence pairs in order to find the POS tag of relevant content words.

\begin{figure}[htbp]
    \centering
    \includegraphics[width=\textwidth]{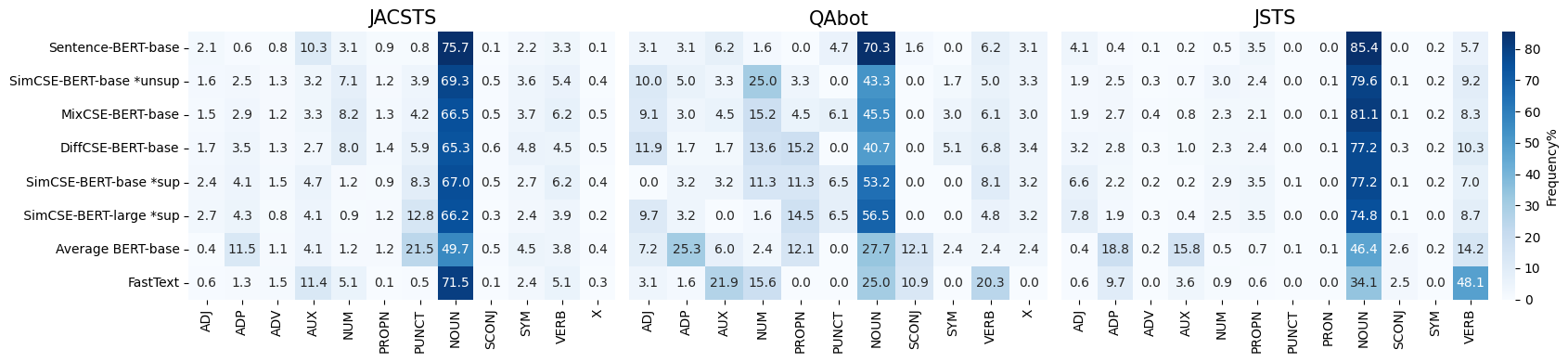}
    \caption{Distributions of the most relevant POS tags for sentence pairs in three datasets. X annotated on the last column represents the set of POS tags differing from the ones annotated on the other columns. *unsup and *sup mean unsupervised and supervised, respectively (see Table~\ref{tab3}).}
    \label{fig3}
\end{figure}

Fig.~\ref{fig3} shows the distributions of $\hat{w}$'s POS tags on three human-annotated datasets, ranging from general domain JSTS dataset and specific domain JACSTS and QABot datasets. For JACSTS and JSTS datasets, semantically similar sentence pairs are collected as the ones that are individually labeled by a semantic similarity score of 4 or 5. For QABot dataset, we pair a query with questions that are labeled as $1$ for this query. In Fig.~\ref{fig3}, one row indicates the distribution of $\hat{w}$'s POS tags by computing sentence embeddings using some sentence embedding models checked in Section~\ref{ssec:j_benchmark_result}. We find that nouns (NOUN) are the POS tag of the most relevant content words for Japanese sentences in all three datasets ranging from multiple domains. This result is important and experimentally validates the theoretical foundation of SDJC. That is, it is practical to substitute nouns in an anchor sentence to construct hard negative sentences.

\subsection{Implementation Details of SDJC}
\label{ssec:implementation_details}

This section provides implementation details of the data generation module and the domain adaption module. 

\subsubsection{Data Generation}
\label{ssec:hard_generation}

The data generator for a specific domain is obtained by fine-tuning the pre-trained T5-base checkpointint\footnote{https://huggingface.co/sonoisa/t5-base-japanese-v1.1} using a specific domain corpus collected in Section~\ref{ssec:specific_domain_corpus_collection}. This fine-tuning is implemented using Pytorch and Huggingface and performed for two epochs where the batch size is $96$ and Adafactor~\cite{shazeer2018adafactor} is employed as an optimizer with a learning rate of $0.001$. The fine-tuning is run on a server equipped with three NVIDIA RTX 2080Ti GPUs. 

The input of the data generator is created by treating each sentence in the specific domain corpus as an anchor sentence and masking all noun chunks in it. The traditional beam search decoding method~\cite{fan2018hierarchical, shao2017generating} implemented in Huggingface~\cite{wolf2019huggingface} is used to generate hard negative sentences. Huggingface also offers beam search decoding methods with probabilistic sampling like beam-search multinomial sampling and nucleus sampling~\cite{holtzman2019curious}. These methods sample the next token from the probability distribution over the entire vocabulary. However, this probabilistic sampling is expected to generate inconsistent or unnatural hard negative sentences. Hence, we opt for the traditional beam search decoding method that simply selects the next token having the overall highest probability for the entire generated tokens.


Table~\ref{tab5} shows two real examples each showing an anchor sentence and hard negative sentences generated from it. The first and second examples are from the clinical and educational domains, respectively. As can be seen from Table~\ref{tab5}, the generated hard negative sentences differ from the anchor sentence in part or all noun chunks. This means that the former have the same syntactic structure but dissimilar semantic meanings to the latter.
Additionally, Table~\ref{tab5} makes us come up with one interesting future work to generate even harder negative sentences. That is, one could introduce subtle synonymous substitutions in part of noun chunks that preserve the overall syntactic structure while creating more nuanced semantic differences. This would encourage a backbone model to learn fine-grained distinctions, improving its robustness in identifying true semantic similarity.


\begin{CJK}{UTF8}{min} 
\begin{table}[htbp]
\centering
\begin{adjustbox}{width=\textwidth}
\begin{tabular}{lllllllllllllll}
\textbf{Clinical Domain}            &            &            &           &          &          &         &         &        &        &        &        &        &        &  \\ \cmidrule{1-14} 
\multicolumn{8}{l}{\textbf{Anchor sentence}}                                                                      &        &        &        &        &        &        &  \\
\multicolumn{8}{l}{幻聴が増悪し、アルコール性精神障害の合併が疑われ、精神科受診が適切と判断された。}                                                        &        &        &        &        &        &        &  \\
\multicolumn{14}{l}{The auditory hallucinations worsened, complications of alcoholic psychosis were suspected, and a psychiatric consultation was deemed appropriate.}     &  \\
\multicolumn{8}{l}{\textbf{Hard negative sentences}}                                                       &        &        &        &        &        &        &  \\
\multicolumn{9}{l}{徐々に症状が増悪し、急性大動脈解離の急性大動脈解離が疑われ、緊急手術が適切と判断された。}                                                             &        &        &        &        &        &  \\
\multicolumn{15}{l}{Gradual worsening   of symptoms, acute aortic dissection of acute aortic dissection was   suspected, and emergency surgery was deemed appropriate.}      \\
\multicolumn{15}{l}{アルコール依存症の症状が増悪し、急性大動脈解離のリスクファクターの増大が疑われ、緊急手術が適切と判断された。}                                                                                                  \\
\multicolumn{14}{l}{The patient's   symptoms of alcoholism were exacerbated and an increased risk factor for   acute aortic dissection was suspected, and}                &  \\
\multicolumn{4}{l}{emergency   surgery was deemed appropriate.}           &          &          &         &         &        &        &        &        &        &        &  \\
\multicolumn{14}{l}{少しずつ症状が増悪し、カテコラミン心筋症の再発や悪化が疑われ、精神科の受診が適切と判断された。}                                                                                                      &  \\
\multicolumn{14}{l}{The symptoms   gradually worsened, and a psychiatric consultation was deemed appropriate due   to a suspected recurrence or}                          &  \\
\multicolumn{4}{l}{worsening   of catecholaminergic cardiomyopathy.}      &          &          &         &         &        &        &        &        &        &        &  \\
\multicolumn{12}{l}{徐々に意識障害が増悪し、急性硬膜外血腫の再発や急性大動脈解離が疑われ、脳底動脈造影CT撮影が適切と判断された。}                                                                            &        &        &  \\
\multicolumn{14}{l}{Gradually worsening   loss of consciousness, recurrent acute epidural hematoma and acute aortic   dissection were suspected, and CT imaging with}     &  \\
\multicolumn{7}{l}{contrast-enhanced   cerebral basilar artery angiography was deemed appropriate.}       &         &        &        &        &        &        &        &  \\
\multicolumn{12}{l}{来院後徐々に症状が増悪し、抗ヒスタゾールや精神科受診の妥当性も低いことが疑われ、精神科受診が適切と判断された。}                                                                            &        &        &  \\
\multicolumn{14}{l}{Symptoms gradually   worsened after the patient's visit, and it was suspected that antihistazol   and psychiatric consultation were not appropriate.} &  \\
\multicolumn{12}{l}{症例は徐々に症状が増悪し、COVID-19の急性増悪のリスクファクターの増悪が疑われ、COVID-19の診断が適切と判断された。}                                                                   &        &        &  \\
\multicolumn{14}{l}{The patient had a   gradual exacerbation of symptoms, and a diagnosis of COVID-19 was deemed   appropriate due to the suspected exacerbation of}      &  \\
\multicolumn{4}{l}{risk factors   for acute exacerbation of COVID-19.}    &          &          &         &         &        &        &        &        &        &        &  \\ 
       &            &            &           &          &          &         &         &        &        &        &        &        &        &  \\
\textbf{Educational Domain}                &            &            &           &          &          &         &         &        &        &        &        &        &        &  \\ \cmidrule{1-14} 
\textbf{Anchor sentence}                  &            &            &           &          &          &         &         &        &        &        &        &        &        &  \\
\multicolumn{14}{l}{とあるクラスで作成されたオブジェクトを他クラスで操作するためにはどのようにすれば良いでしょうか?}                                                                                                     &  \\
\multicolumn{14}{l}{How can an object   created in one class be manipulated by another class?}                                                                            &  \\
\multicolumn{3}{l}{\textbf{Hard negative sentence}}        &           &          &          &         &         &        &        &        &        &        &        &  \\
\multicolumn{14}{l}{MiniDuckSimulator.javaで作成されたオブジェクトをSystem.javaで操作するためにはどのようにすれば良いでしょうか?}                                                                              &  \\
\multicolumn{14}{l}{How can I use   System.java to manipulate objects created in MiniDuckSimulator.java?}                                                                 &  \\
\multicolumn{14}{l}{AdderServletクラスで作成されたオブジェクトをServletクラスで操作するためにはどのようにすれば良いでしょうか?}                                                                                      &  \\
\multicolumn{13}{l}{How can I   manipulate objects created in the AdderServlet class with the Servlet class?}                                                    &        &  \\
\multicolumn{13}{l}{MetroFair.javaで作成されたオブジェクトをSafariの環境で操作するためにはどのようにすれば良いでしょうか?}                                                                               &        &  \\
\multicolumn{13}{l}{How can I   manipulate objects created in MetroFair.java in the Safari environment?}                                                         &        &  \\
\multicolumn{13}{l}{MallardDuckクラスで作成されたオブジェクトをSystem.javaで操作するためにはどのようにすれば良いでしょうか?}                                                                             &        &  \\
\multicolumn{13}{l}{How can I use   System.java to manipulate objects created in the MallardDuck class?}                                                         &        &  \\
\multicolumn{13}{l}{DecoyDuckクラスで作成されたオブジェクトをGameDuckクラスで操作するためにはどのようにすれば良いでしょうか?}                                                                               &        &  \\
\multicolumn{13}{l}{How can an object   created in the DecoyDuck class be manipulated in the GameDuck class?}                                                    &        &  \\
\multicolumn{13}{l}{DecoyDuckSimulator.javaで作成されたオブジェクトをGecoyDuckSimulator.javaで操作するためにはどのようにすれば良いでしょうか?}                                                        &        &  \\
\multicolumn{14}{l}{How can I use   GecoyDuckSimulator.java to manipulate objects created in   DecoyDuckSimulator.java?}                                                  &
\end{tabular}
\end{adjustbox}
\caption{Examples each of which shows an anchor sentence and hard negative sentences generated from it in the clinical or education domains.} 
\label{tab5}
\end{table}
\end{CJK}


\subsubsection{Domain Adaptation}


This module is implemented using Pytorch and Huggingface. We test different backbone models that are defined by public Japanese BERT checkpoints\footnote{https://huggingface.co/tohoku-nlp}. Each of these BERT models is used to extract the embedding of a sentence as the average embeddings of the tokens on its top layer. AdamW optimizer~\cite{loshchilov2017decoupled} is used by setting the learning rate to $5e^{-5}$ and $1e^{-5}$ for BERT-base and BERT-large, respectively. The batch size is $512$ for all models. Also, hyperparameter tuning is performed to find adequate values of the temperature $\tau$ and the weight for discriminating hard negative pairs $\alpha$ by setting different values to them in the training setting. In addition, for performance improvement, we checked the volumes of training data and decided to adapt BERT-base by generating four and five hard negative sentences for every anchor sentence in the clinical and educational domains, respectively. Similarly, the adaptation of BERT-large is done by generating five and six hard negative sentences in the clinical and educational domains, respectively. We saved every adapted backbone model at different training steps, tested it on the domain-specific downstream task and published the best one on our github repository\footnote{https://github.com/ccilab-doshisha/SDJC}. The above-mentioned domain adaptation is run on a server with a single NVIDIA A100 GPU.


\subsection{Main Results}
\label{ssec:main_results}

Like Section~\ref{ssec:j_benchmark_result}, we use the term, sentence embedding model, to indicate a combination of a sentence representation method and a backbone model. To examine the effectiveness of SDJC's domain adaption, the sentence embedding models listed in the fourth column of Table~\ref{tab7} are compared on the two domain-specific downstream tasks described in Section~\ref{ssec:downstream_tasks}. The first to third columns in Table~\ref{tab7} depict the categorization of these models into the following seven groups:

\begin{sidewaystable}
\centering
\begin{tabular}{llcclccccc}
\hline
\multirow{2}{*}{\shortstack[c]{\textbf{Adaptation}\\ \textbf{type}}} & \multirow{2}{*}{\shortstack[c]{\textbf{Embedding}\\ \textbf{type}}} & \multirow{2}{*}{\shortstack[c]{\textbf{Fine-tuning}\\ \textbf{data}}} & \textbf{Group} & \multirow{2}{*}{\shortstack[c]{\textbf{Sentence embedding}\\ \textbf{model}}} & \textbf{JACSTS}&\multicolumn{4}{c}{\textbf{QABot}} \\ \cmidrule{7-10}
& & & &  &   & \textbf{MRR}    & \textbf{MAP}    & \textbf{P@1}  & \textbf{P@5}  \\ \hline \hline
\multirow{19}{*}{\shortstack[c]{No domain\\ adaptation}} & \multirow{6}{*}{\shortstack[c]{Word-view}} & \multirow{6}{*}{None} & \multirow{6}{*}{1} & FastText embedding & 75.26 & 0.5666 & 0.4153 & 0.45 & 0.22 \\
& & & & BERT-base & 76.73 & 0.6931 & 0.5206 & 0.6 & 0.24 \\ 
& & & & BERT-base-v2 &77.55 &0.5307 &0.3886 &0.35 &0.22\\
& & & & BERT-base-v3 &78.09 &0.6098 &0.4736 &0.55 &0.2 \\
& & & & BERT-large &68.09 &0.5468 &0.4244 &0.45 &0.2 \\
& & & & BERT-large-v2 &74.56 &0.5864 &0.4893 &0.5 &0.25 \\ \cmidrule{2-10}
& \multirow{13}{*}{\shortstack[c]{Sentence-view}} & \multirow{8}{*}{\shortstack[c]{\textit{wiki1m}\\(general\\domain)}} & \multirow{8}{*}{2} & SimCSE-BERT-base-v3 & 79.92 & 0.7645 & 0.6121 & 0.7 & 0.31 \\
& & & & SimCSE-BERT-large  &81.10 &0.7304 &0.6372 &0.65 &0.32\\
& & & & SimCSE-BERT-large-v2 &79.43 &\textbf{0.9102} &\textbf{0.7573} &\textbf{0.9} &0.33 \\
& & & & MixCSE-BERT-base &80.94 &0.7591 &0.6261 &0.7 &0.32 \\
& & & & MixCSE-BERT-base-v3 & 78.66 & 0.7935 & 0.6564 & 0.75 & 0.31 \\
& & & & MixCSE-BERT-large &\textbf{81.71} &0.8468 &0.7188 &0.8 &0.34 \\
& & & & MixCSE-BERT-large-v2 &79.65 &0.8876 &0.7280 &0.85 &\textbf{0.35} \\
& & & & DiffCSE-BERT-base & 79.82 & 0.7259 & 0.6179 & 0.65 & 0.32 \\ \cmidrule{3-10}
& & \multirow{5}{*}{\shortstack[c]{JSNLI\\(general\\domain)}} & \multirow{5}{*}{3} & Sentence-BERT-base & 75.69 & 0.5922 & 0.4080 & 0.55 & 0.21 \\
& & & & Sentence-BERT-large  & 73.77 & 0.5263 & 0.3475 & 0.45 & 0.19 \\
& & & & SimCSE-BERT-base-v3  & 81.52 & 0.7207 & 0.5933 & 0.6  & 0.3 \\
& & & & SimCSE-BERT-large & 81.67 &0.7664 &0.6208 &0.7 &0.32\\
& & & & SimCSE-BERT-large-v2 &79.61 &0.7863 &0.5971 &0.75 &0.3 \\ \hline \hline
\multirow{6}{*}{\shortstack[c]{Domain\\ adaptation}} & \multirow{6}{*}{\shortstack[c]{Sentence-view}} & \multirow{3}{*}{In-domain} & \multirow{3}{*}{4} & SDJC-base-v3 & \textbf{83.00} & 0.8196 & 0.6572 & \textbf{0.8} & \textbf{0.34} \\
& & & & SDJC-large & 81.63 & 0.8143 & \textit{0.7063} & 0.75 & 0.34 \\
& & & & SDJC-large-v2 & 79.28 & 0.7868 & 0.6340 & 0.75 & 0.34 \\ \cmidrule{3-10}
& & \multirow{3}{*}{ \shortstack[c]{\textit{wiki1m} \ \& \\ Specific domain corpus} } & \multirow{3}{*}{5} & $\overline{\mathrm{SDJC}}$-base-v3 & 82.61 & 0.8093 & 0.6489 & 0.75 & 0.34 \\
& & & & $\overline{\mathrm{SDJC}}$-large & 82.05 &0.8368 &0.6771 &0.8 &0.33 \\
& & & & $\overline{\mathrm{SDJC}}$-large-v2 & 80.99 &\textbf{0.8476} &0.6766 &0.8 &0.33 \\ \hline \hline
\multirow{3}{*}{\shortstack[c]{Two-stage domain\\ adaptation}} & \multirow{3}{*}{\shortstack[c]{Sentence-view}} & \multirow{2}{*}{In-domain$\rightarrow$JSNLI} & \multirow{2}{*}{6} & SDJC-base-v3-JSNLI & \textbf{83.97} & 0.8690 & 0.7045 & 0.85 & 0.34 \\
& & & & SDJC-large-JSNLI & 82.60  & 0.8516 & 0.7490 &0.8 &\textbf{0.37} \\ \cmidrule{3-10} 
& & In-domain \ $\rightarrow$ \ \textit{wiki1m} & 7 & SDJC-large-v2-wiki  &80.56 &\textbf{0.9176} &\textbf{0.7860} &\textbf{0.9} &0.35\\
\hline
\end{tabular}
\caption{Performance on two domain-specific downstream tasks.}
\label{tab7}
\end{sidewaystable}

\noindent \textbf{Group 1:} 
The sentence embedding models in this group are very simple and serve as general baselines in the sense that they use neither domain adaptation nor sentence-view embedding. In other words, these models are pre-trained on general domain datasets to compute embeddings of tokens, not sentences. The embedding of a sentence is computed by directly averaging embeddings of tokens in it.

\noindent \textbf{Group 2:} This group includes sentence embedding models that use unsupervised contrastive learning methods on a general domain dataset, especially \textit{wiki1m} described in Section~\ref{ssec:j_benchmark_result}. These models are treated as unsupervised baselines in the sense that they adopt sentence-view embeddings with no domain adaptation. 




\noindent \textbf{Group 3:} Compared to Group 2, this group includes sentence embedding models that employ supervised sentence representation methods on a general domain labeled dataset (i.e., JSNLI). These models are regarded as supervised baselines that use sentence-view embeddings with no domain adaptation.


\noindent \textbf{Group 4:} This group includes three sentence embedding models obtained by adapting different backbone models with our  proposed SDJC. It should be noted that a dataset consisting of a specific domain corpus and hard negative sentences generated by SDJC's data generator is called an \textit{in-domain dataset}. 

  
\noindent \textbf{Group 5:} The sentence embedding models in this group are trained in an unsupervised style using an unlabeled dataset that consists of \textit{wiki1m} and a specific domain corpus. We compare this group with Group 4 to examine the necessity of generating hard negative sentences.


\noindent \textbf{Group 6:} This group includes the sentence embedding models that are obtained by firstly adapting backbone models with SDJC and further fine-tuning them on labeled JSNLI dataset. This is a supplemental approach to test a hypothesis that the second fine-tuning with a general domain dataset may be useful for regularizing sentence embedding models adapted by SDJC. This kind of two-stage domain adaptation is adopted in existing studies~\cite{gao2021simcse, chen2022generate}.


\noindent \textbf{Group 7:} This group only includes one sentence embedding model, SDJC-large-v2-wiki, that is trained by the same two-stage domain adaptation as Group 6. But, instead of labeled JSNLI dataset, the second stage uses unlabeled \textit{wiki1m} dataset to perform unsupervised contrastive learning described in Group 2. 

Table~\ref{tab7} shows that SDJC improves the performance through domain adaptation using an in-domain dataset. Specifically, by comparing Group 4 with Groups 2 and 3, SDJC can achieve competitive results against sentence embedding models fine-tuned on general domain datasets, although SDJC uses an in-domain dataset whose scale is just $10$\% of the datasets used in Groups 2 and 3. This validates the effectiveness of SDJC for building sentence embedding models in low-resource languages. Also, Groups 6 and 7 indicate that sentence embedding models obtained by SDJC can be improved by the additional fine-tuning with a general domain dataset. Surprisingly, we find from Group 2 that the convincing performance on QABot dataset in the educational domain is achieved by SimCSE-BERT-large-v2, which is obtained by fine-tuning BERT-large-v2 on \textit{wiki1m} with no domain adaptation. As seen from Group 7, this performance is further improved by adding SDJC's domain adaptation before fine-tuning on \textit{wiki1m}.
Moreover, SDJC-base-v3 offers better or comparable performances compared to the sentence embedding models in Group 5, which indicates the effectiveness of hard negative sentences generated by SDJC's data generator. Overall, SDJC achieves powerful performances with a small-scale dataset synthesized with hard negative sentences, and gets the best performances on almost all metrics after the additional fine-tuning.

\subsection{Visualization of Sentence Embeddings}
\label{ssec:visualization_se}

\begin{figure}[htbp]
    \centering
    \includegraphics[width=\textwidth]{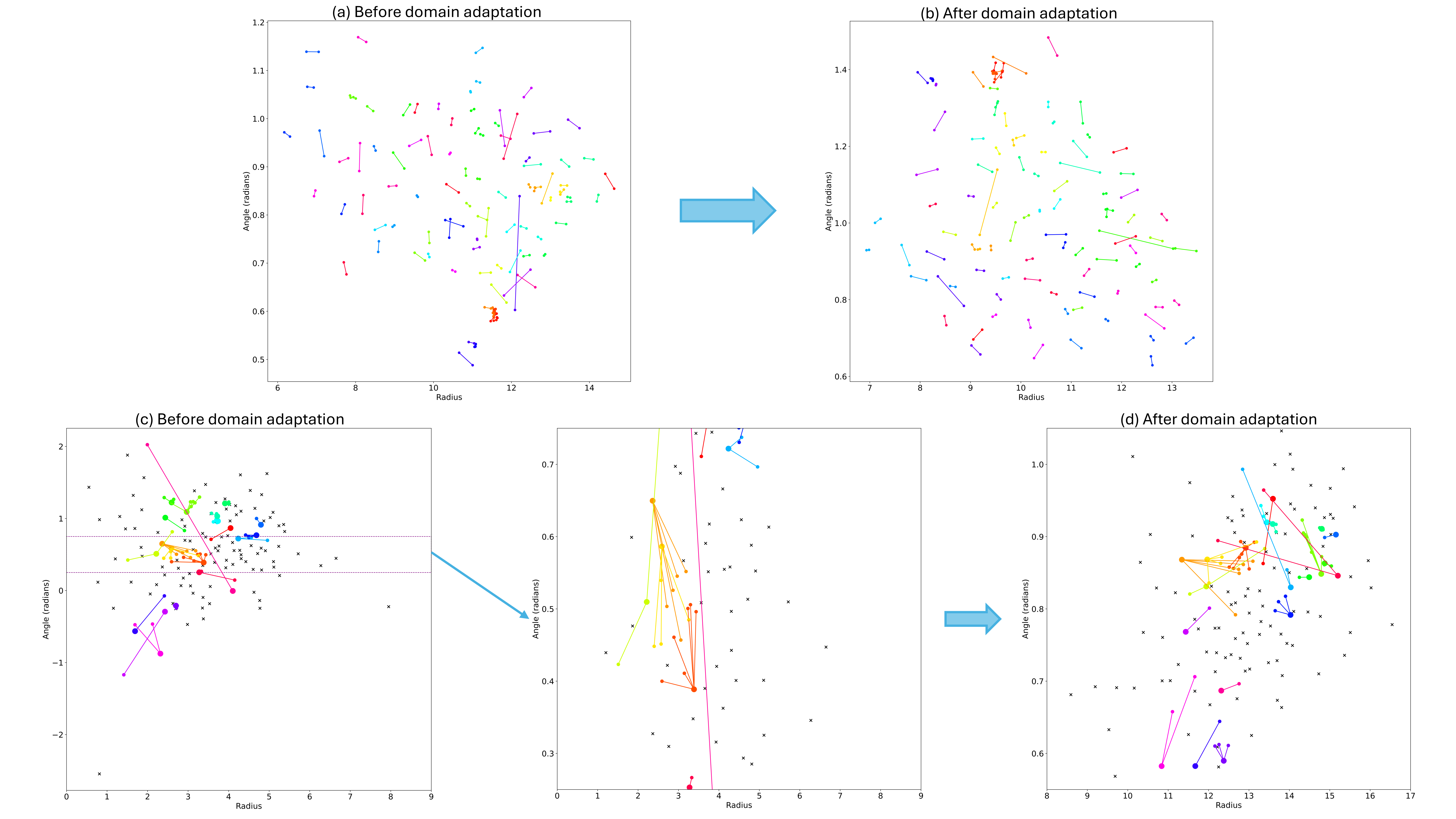}
    \caption{Visual comparisons between sentence embeddings obtained by a backbone model before and after SDJC's domain adaptation.}
    \label{fig_embedding_visual}
\end{figure}

To visually check that SDJC produces improved sentence embeddings, we visualize and compare those embeddings to the ones before SDJC's domain adaptation. 
Fig.~\ref{fig_embedding_visual} shows two visualization pairs. The first pair targets at JACSTS dataset in the clinical domain and shows the comparison between sentence embeddings obtained by SimCSE-BERT-base-v3 in Table~\ref{tab7} (top-left visualization) and the ones by SDJC-base-v3 (top-right visualization). The second pair is for comparing between sentence embeddings by SimCSE-BERT-base-v3 (bottom-left visualization) and the ones by SDJC-base-v3 (bottom-right visualization) for QABot dataset in the educational domain. 
The reason why SimCSE-BERT-base-v3 and SDJC-base-v3 are selected is that the latter is SDJC's default configuration leading to stable performances and the former is the most related to it. 

For the visualizations for JACSTS dataset, we aim to visualize semantically similar sentence pairs and examine whether their locations in a 2D space are close to each other. JACSTS dataset contains more than $1,300$ sentence pairs annotated with scores of $4$ or $5$ (see Figure~\ref{fig4}), so visualizing all of them causes a visibility problem. Hence, we opt to visualize $100$ randomly sampled sentence pairs and share them in both the top-left and top-right visualizations of Fig.~\ref{fig_embedding_visual}. We use UMAP~\cite{mcinnes2020umap} to reduce the dimensionality of sentence embeddings by preserving their cosine distances (i.e., $1 - \mbox{(cosine similarity)}$ in the original space.
Next, we transform the reduced embeddings into polar coordinates where the x-axis represents the length of each embedding and the y-axis represents its angle.
Each pair of sentence embeddings is connected by a line with a distinct color which indicates that they belong to the same sentence pair. Colors for sentence pairs are designed to be evenly distributed in the RGB space. 
In addition, the proximity between the vertical positions (angles) of two embeddings corresponds to their cosine similarity. Thus, we can consider that embeddings are high-quality if sentence pairs are connected with horizontally aligned lines. On the other hand, many vertical aligned lines exhibit low-quality embeddings. It can be seen that semantically similar sentence pairs in the top-right visualization of Fig.~\ref{fig_embedding_visual} are depicted in more horizontally aligned lines compared to the top-left visualization. This means that these sentence pairs become more coherent after SDJC's domain adaptation.

Our purpose for the visualizations for QABot dataset is to analyze the similarities between each of $20$ queries and its relevant questions. For the visibility issue, we decide to visualize $53$ questions that are relevant to one of $20$ queries and $100$ randomly sampled irrelevant questions. These questions as well as $20$ queries are commonly used in both of the bottom-left and bottom-right visualizations of Fig.~\ref{fig_embedding_visual}. Similar to JACSTS dataset, UMAP is use to project their embeddings into a 2D space and then transferred them into a polar coordinate. Each query and its relevant questions are connected with lines of a distinct color, while the other questions are marked by cross $\times$. At first glance, we cannot see a significant difference between the frequency of horizontally aligned lines in the bottom-right visualization and the one in the bottom-left. However, the range of the y-axis in the former is much larger than the one in the latter (the range of the x-axis is the same between them). This means that lines in the bottom-right might be misinterpreted as horizontally aligned. Thus, we provide the bottom-middle visualization that is the enlarged visualization of the embeddings in the purple box of the bottom-left and is characterized by the same range of the y-axis to the bottom-right.
This enlarged visualization clearly shows that lines in the bottom-left are actually vertically aligned and several irrelevant questions are more closer to queries in terms of their vertical positions. Therefore, we can conclude that SDJC's domain adaptation makes embeddings of each query and its relevant questions closer to each other.

\section{Conclusion and Future Work}
\label{sec5}

In this paper, we introduced a novel, effective domain adaptation approach, SDJC, to overcome the scarcity of large-scale labeled datasets for a low-resource language like Japanese. The core of SDJC is the contrastive learning module that takes advantage of hard negative sentences which a data generator generates by replacing noun chunks in anchor sentences. In addition, a sentence representation method and a backbone model adopted in SDJC are selected through a Japanese STS benchmark study. Here, our comprehensive JSTS benchmark dataset is established to evaluate various existing sentence representation methods and backbone models for Japanese. Finally, the effectiveness of SDJC's domain adaptation is demonstrated on the two domain-specific downstream tasks, the STS task in a clinical domain and the information retrieval task in an educational domain. On \url{https://github.com/ccilab-doshisha/SDJC}, all the datasets, codes and models adapted by SDJC are available except the educational domain datasets that concern the ethical issue. We believe that SDJC paves a practical way for domain-specific downstream tasks in low-resource languages.

In the future, we plan to explore a strategy to generate higher-quality hard negative sentences by selectively replacing some of noun chunks in an anchor sentence or utilizing an LLM. In addition, we will evaluate SDJC on more diverse domain-specific downstream tasks.


\backmatter


\bmhead{Data availability}

Except the educational domain datasets that concern the ethical issue, all the datasets, codes and models adapted by SDJC are available online: \url{https://github.com/ccilab-doshisha/SDJC}.

\bmhead{Conflict of interest}

All authors declare that they have no known competing financial interests or personal relationships that could have appeared to influence the work reported in this paper.

\begin{appendices}

\end{appendices}


\bibliography{references_v2}

\end{document}